\documentclass[10pt,letterpaper]{article}
\usepackage[top=0.85in,left=2.75in,footskip=0.75in]{geometry}

\usepackage{amsmath,amssymb}

\usepackage{changepage}

\usepackage[utf8x]{inputenc}

\usepackage{textcomp,marvosym}

\usepackage{cite}

\usepackage{nameref}

\usepackage{microtype}
\DisableLigatures[f]{encoding = *, family = * }

\usepackage[table]{xcolor}

\usepackage{array}

\newcolumntype{+}{!{\vrule width 2pt}}

\newlength\savedwidth

\raggedright
\usepackage[aboveskip=1pt,labelfont=bf,labelsep=period,justification=raggedright,singlelinecheck=off]{caption}

\makeatletter
\renewcommand{\@biblabel}[1]{\quad#1.}
\makeatother

\usepackage{lastpage,fancyhdr,graphicx}
\usepackage{epstopdf}
\fancyheadoffset[L]{2.25in}
\fancyfootoffset[L]{2.25in}
\usepackage{graphicx}
\usepackage{caption}
\usepackage{subcaption}
\usepackage{gensymb}
\usepackage{multirow}
\usepackage{booktabs}
\usepackage{rotating}
\usepackage{tabularx}
\usepackage{hyphenat}
\usepackage{makecell}

\usepackage[]{siunitx}
\usepackage{textcomp}
\usepackage{pdflscape}
\usepackage{afterpage}

\begin{document}
\vspace*{0.2in}

\begin{flushleft}
{\Large
\textbf\newline{Deep learning approach to describing and classifying fungi microscopic images} 
}
\newline
\\
Bartosz Zieli\'nski\textsuperscript{1,2*},
Agnieszka Sroka-Oleksiak\textsuperscript{3,4},
Dawid Rymarczyk\textsuperscript{1,2},
Adam Piekarczyk\textsuperscript{1},
Monika Brzychczy-W\l{}och\textsuperscript{4}
\\
\bigskip
\textbf{1} Faculty of Mathematics and Computer Science, Jagiellonian University, 6~\L{}ojasiewicza Street, 30-348 Krak\'ow, Poland
\\
\textbf{2} Ardigen, 76~Podole Street, 30-394 Krak\'ow, Poland
\\
\textbf{3} Department of Mycology, Chair of Microbiology, Faculty of Medicine, Jagiellonian University Medical College, 18~Czysta Street, 31-121 Krak\'ow, Poland
\\
\textbf{4} Department of Molecular Medical Microbiology, Chair of Microbiology, Faculty of Medicine, Jagiellonian University Medical College, 18~Czysta Street, 31-121 Krak\'ow, Poland
\\
\bigskip

* bartosz.zielinski@uj.edu.pl

\end{flushleft}
\section*{Abstract}
Preliminary diagnosis of fungal infections can rely on microscopic examination. However, in many cases, it does not allow unambiguous identification of the species due to their visual similarity. Therefore, it is usually necessary to use additional biochemical tests. That involves additional costs and extends the identification process up to 10 days. Such a delay in the implementation of targeted therapy may be grave in consequence as the mortality rate for immunosuppressed patients is high. In this paper, we apply a machine learning approach based on deep neural networks and bag-of-words to classify microscopic images of various fungi species. Our approach makes the last stage of biochemical identification redundant, shortening the identification process by 2-3 days, and reducing the cost of the diagnosis.

\section*{Introduction}

\noindent Yeast and yeast-like fungi are a component of natural human microbiota~\cite{Arendrup2013CandidaAC}. However, as opportunistic pathogens, they can cause surface and systemic infections~\cite{CandidaGlabrata}. The leading causes of the fungal infections are impaired function of the immune system and imbalanced microbiota composition in the human body. Other factors of fungal infections include steroid treatment, invasive medical procedures, and long-term antibiotic treatment with a broad spectrum of antimicrobial agents~\cite{CandidaParapsilosis,CandidaX3,Silveira}.

\begin{figure}
    \centering
    \includegraphics[width=\textwidth]{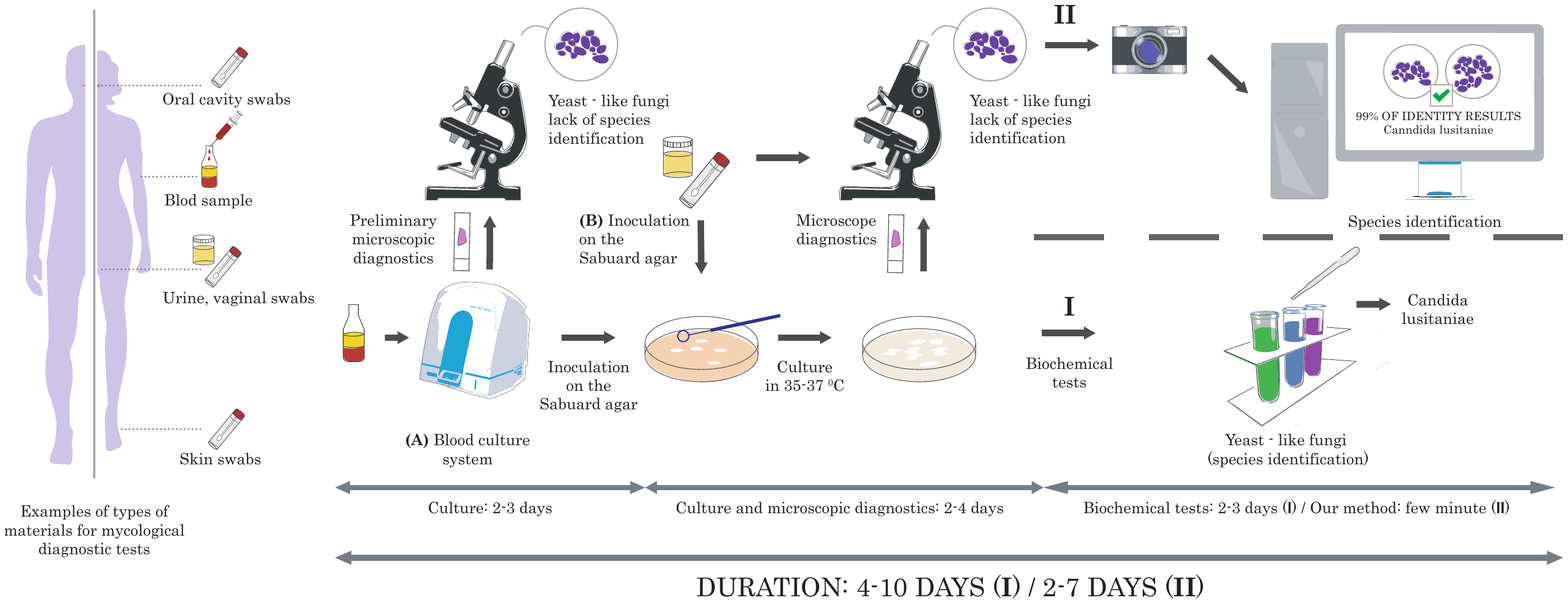}
    \caption{\footnotesize Standard mycological diagnostics (I) require analysis with biochemical tests. As a result, the entire diagnostic process can last 4-10 days. In our computer-aided approach (II), biochemical tests are replaced with a machine learning approach that predicts fungi species based only on microscopic images. It shortens the diagnosis by 2-3 days.}
    \label{fig:schematy}
\end{figure}

The standard procedure in mycological diagnostics begins with collecting various types of test materials like swabs, scraps of skin lesions, urine, blood, or cerebrospinal fluid. Next, the clinical materials (marked as \textbf{B} in Fig.~\ref{fig:schematy}) are directly cultured on special media, while the blood and cerebrospinal fluid samples (marked as \textbf{A} in Fig.~\ref{fig:schematy}) require prior cultivation in automated closed systems for additional 2-3 days. Material incubates under specific temperature conditions (usually for 2-4 days in case of yeast-like fungi). The initial identification of fungi bases on the assessment of the cells' shapes observed under the microscope as well as the growth rate, type, shape, color, and the smell of the colonies. Such analysis allows the assignment to fungi type; however, identification of the species is usually impossible due to the significant similarity between them. Because of that, further analysis consisting of biochemical tests, is necessary. As a result, the entire diagnostic process from the moment of culture to species identification can last 4-10 days (see Fig.~\ref{fig:schematy}).

In this paper, we apply a machine learning approach based on deep neural networks and bag-of-words approaches to classify microscopic images of various fungus species. As a result, the last stage of biochemical identification is unnecessary, which shortens the identification process by 2-3 days and reduces the cost of diagnosis. It allows accelerating the decision about the introduction of an appropriate antifungal drug, which prevents the progression of the disease and shortens the time of patient recovery.

According to our best knowledge, there are no other methods for classifying fungi species based only on microscopic images. Existing methods involve techniques such as morphological identification of a type of fungi~\cite{papagianni2014characterization}, fluorescence in situ hybridization (FISH)~\cite{lakner2012evaluation}, biochemical techniques, molecular approaches, such as PCR~\cite{ferrer2001detection}, and sequencing~\cite{raja2017fungal}. However, all of them are costly. On the other hand, our method bases on basic microbiological staining (Gram staining) and a simple microscope equipped with a camera, and takes only a few minutes, which makes it easily applicable in many laboratories.

The paper is structured as follows. First, we introduce a fungus database and describe a classification method based on deep neural networks and bag-of-words methods. Then, we present experimental setup, results, and conclusion.

\section*{Materials and methods}
\label{sec:materialsMethods}

\textbf{Materials.} One of the most common fungal infections is candidiasis~\cite{Silveira}, mainly caused by \textit{Candida albicans} (50-70\% of cases)~\cite{Sensitization}. Other species responsible for the diseases are \textit{Candida glabrata}~\cite{CandidaGlabrata,CandidaParapsilosis}, \textit{Candida tropicalis}~\cite{CandidaX3}, \textit{Candida krusei}~\cite{CandidaKrusei}, and \textit{Candida parapsilosis}~\cite{CandidaParapsilosis,CandidaX3}. In high-risk patients, severe infections can also be caused by \textit{Cryptococcus neoformans}~\cite{Saha2} and \textit{Saccharomyces phylum}~\cite{Invasive}. Taking those facts into consideration, we prepared database, which consists of five yeast-like fungal strains: \textit{Candida albicans} ATCC 10231 (CA), \textit{Candida glabrata} ATCC 15545 (CG), \textit{Candida tropicalis} ATCC 1369 (CT), \textit{Candida parapsilosis} ATCC 34136 (CP), and \textit{Candida lustianiae} ATCC 42720 (CL); two yeast strains: \textit{Saccharomyces cerevisae} ATCC 4098 (SC) and \textit{Saccharomyces boulardii} ATCC 74012 (SB); and two strains belonging to the Basidiomycetes: \textit{Maalasezia furfur} ATCC 14521 (MF) and \textit{Cryptococcus neoformans} ATCC 204092 (CN). All strains are from the American Type Culture Collection. The species in our database highly overlap with the most common fungal infections; however, they are not identical due to the limitations of our repository.

The strains were cultured on Sabouraud agar at $37\degree$C for 48h (together with olive oil in the case of Maalaseizia furfur). After this time, microscopic preparations were made (2 preparations for each fungal strain) and stained with Gram method. Images were taken using an Olympus BX43 microscope with 100 times a super-apochromatic objective under oil-immersion. The photographic documentation was then produced with an Olympus BP74 camera and CellSense software (Olympus).

\begin{figure}
    \centering
    \begin{subfigure}[t]{0.49\textwidth}
    \includegraphics[width=\textwidth]{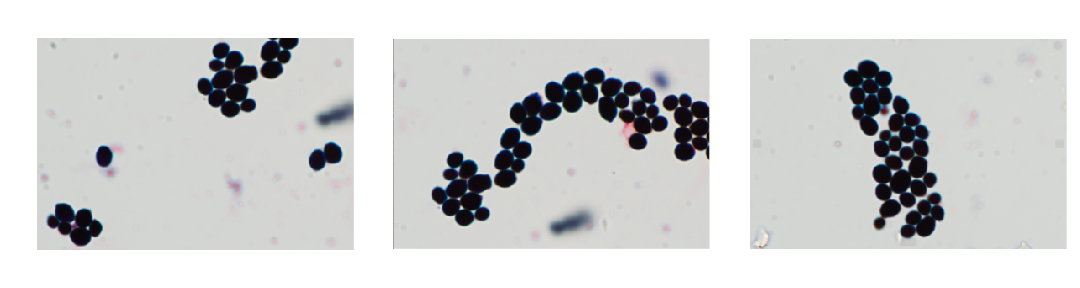}
    \caption{\footnotesize \centering \textit{Candida albicans} (CA)}
    \end{subfigure}
    \vspace{1em}
    \begin{subfigure}[t]{0.49\textwidth}
    \includegraphics[width=\textwidth]{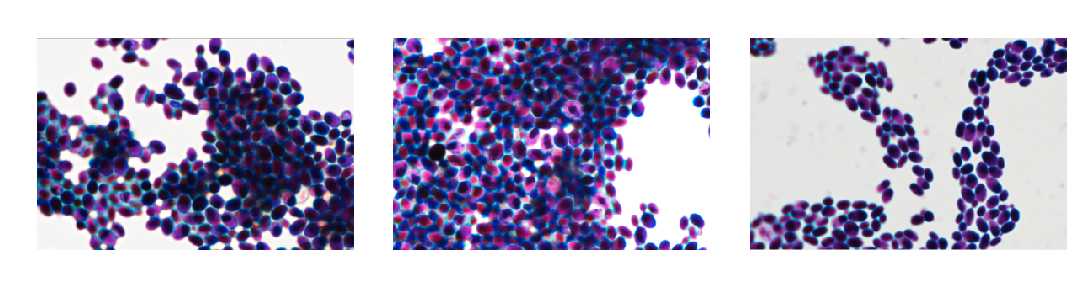}
    \caption{\footnotesize \centering \textit{Candida glabrata} (CG)}
    \end{subfigure}
    \begin{subfigure}[t]{0.49\textwidth}
    \includegraphics[width=\textwidth]{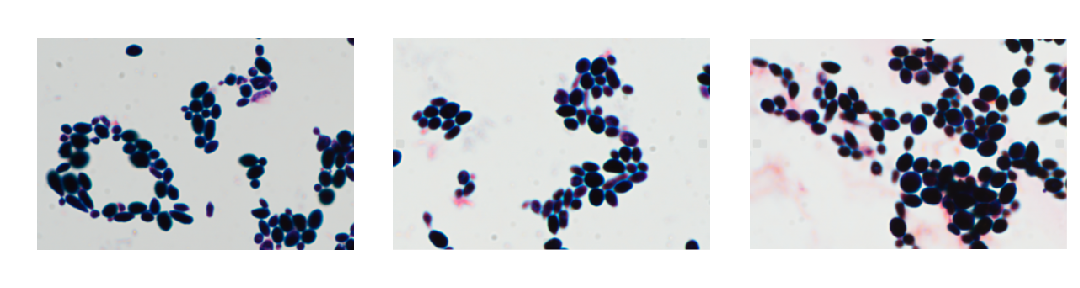}
    \caption{\footnotesize \centering \textit{Candida lustianiae} (CL)}
    \end{subfigure}
    \begin{subfigure}[t]{0.49\textwidth}
    \includegraphics[width=\textwidth]{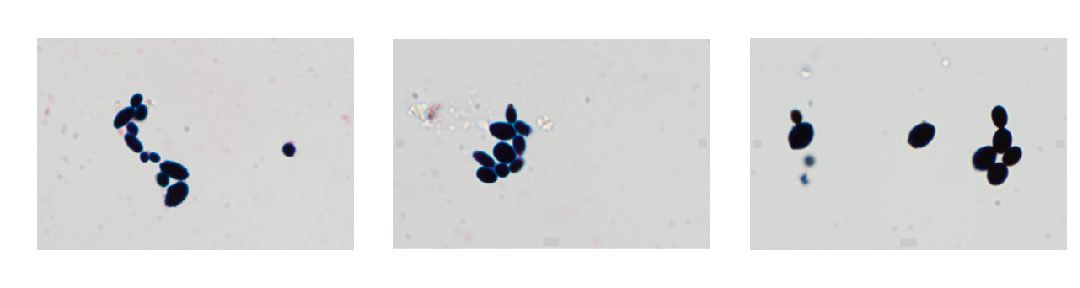}
    \caption{\footnotesize \centering \textit{Cryptococcus neoformans} (CN)}
    \end{subfigure}
    \begin{subfigure}[t]{0.49\textwidth}
    \includegraphics[width=\textwidth]{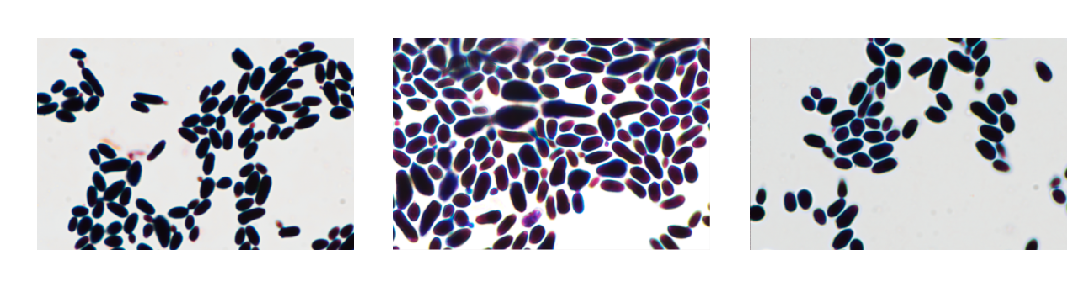}
    \caption{\footnotesize \centering \textit{Candida parapsilosis} (CP)}
    \end{subfigure}
    \begin{subfigure}[t]{0.49\textwidth}
    \includegraphics[width=\textwidth]{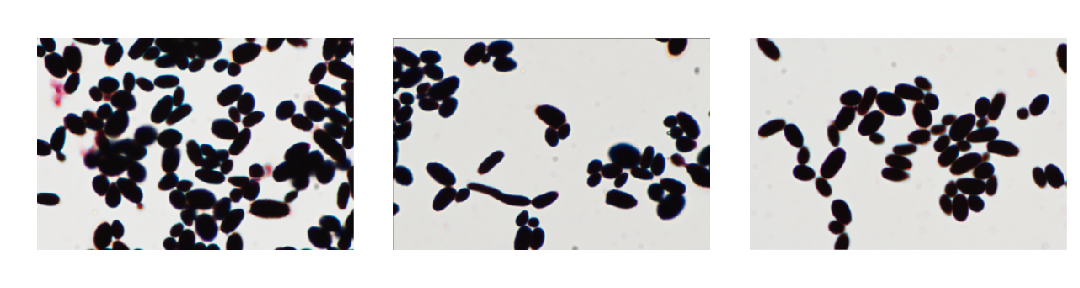}
    \caption{\footnotesize \centering \textit{Candida tropicalis} (CT)}
    \end{subfigure}
    \begin{subfigure}[t]{0.49\textwidth}
    \includegraphics[width=\textwidth]{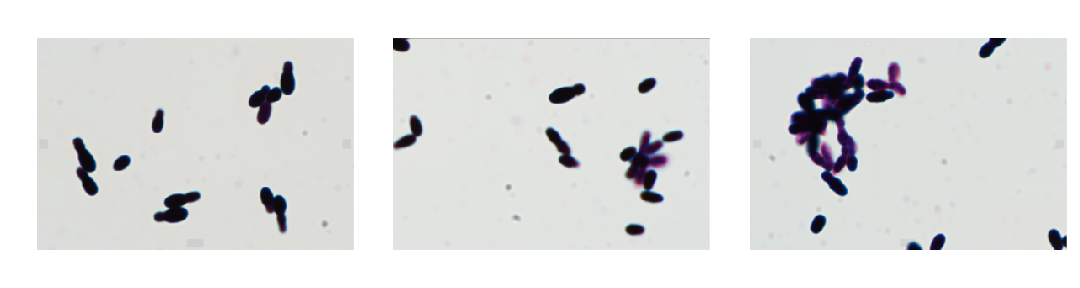}
    \caption{\footnotesize \centering \textit{Maalasezia furfur} (MF)}
    \end{subfigure}
    \begin{subfigure}[t]{0.49\textwidth}
    \includegraphics[width=\textwidth]{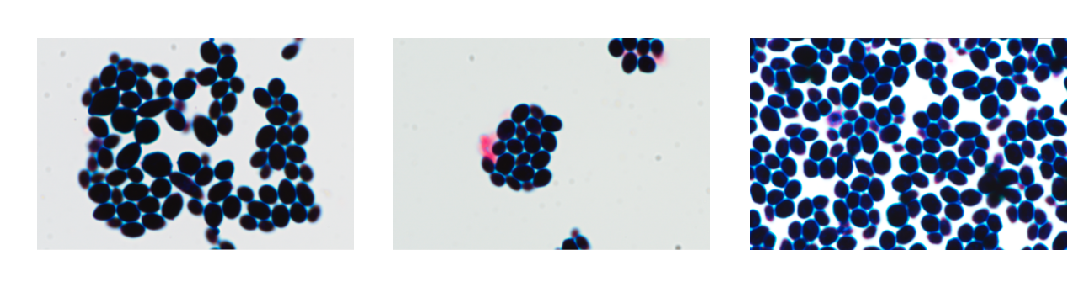}
    \caption{\footnotesize \centering \textit{Saccharomyces boulardii} (SB)}
    \end{subfigure}
    \begin{subfigure}[t]{0.49\textwidth}
    \includegraphics[width=\textwidth]{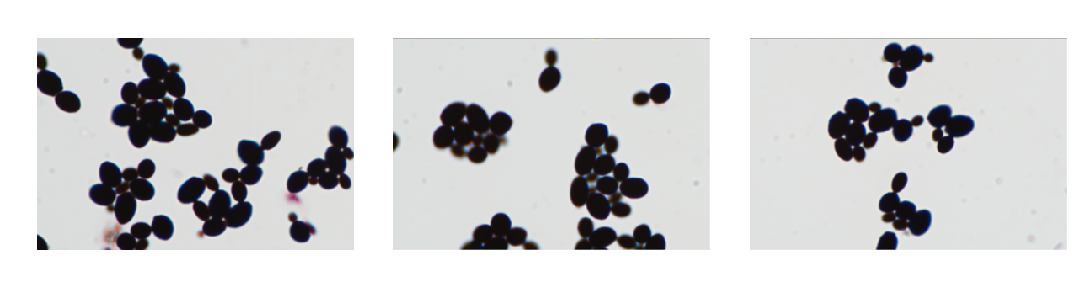}
    \caption{\footnotesize \centering \textit{Saccharomyces cerevisae} (SC)}
    \end{subfigure}
    \caption{Three random images for each of the strains from our DIFaS database.}
    \label{fig:funguses}
\end{figure}

Altogether, our Digital Images of Fungus Species database (DIFaS) contains 180 images (9 strains $\times$ 2 preparations $\times$ 10 images) of resolution $3600\times 5760\times 3$ with 16-bits intensity range in every pixel. In~Fig.~\ref{fig:funguses}, we present three random thumbnails for each of the registered strains.

\begin{figure}
    \centering
    \includegraphics[width=\textwidth]{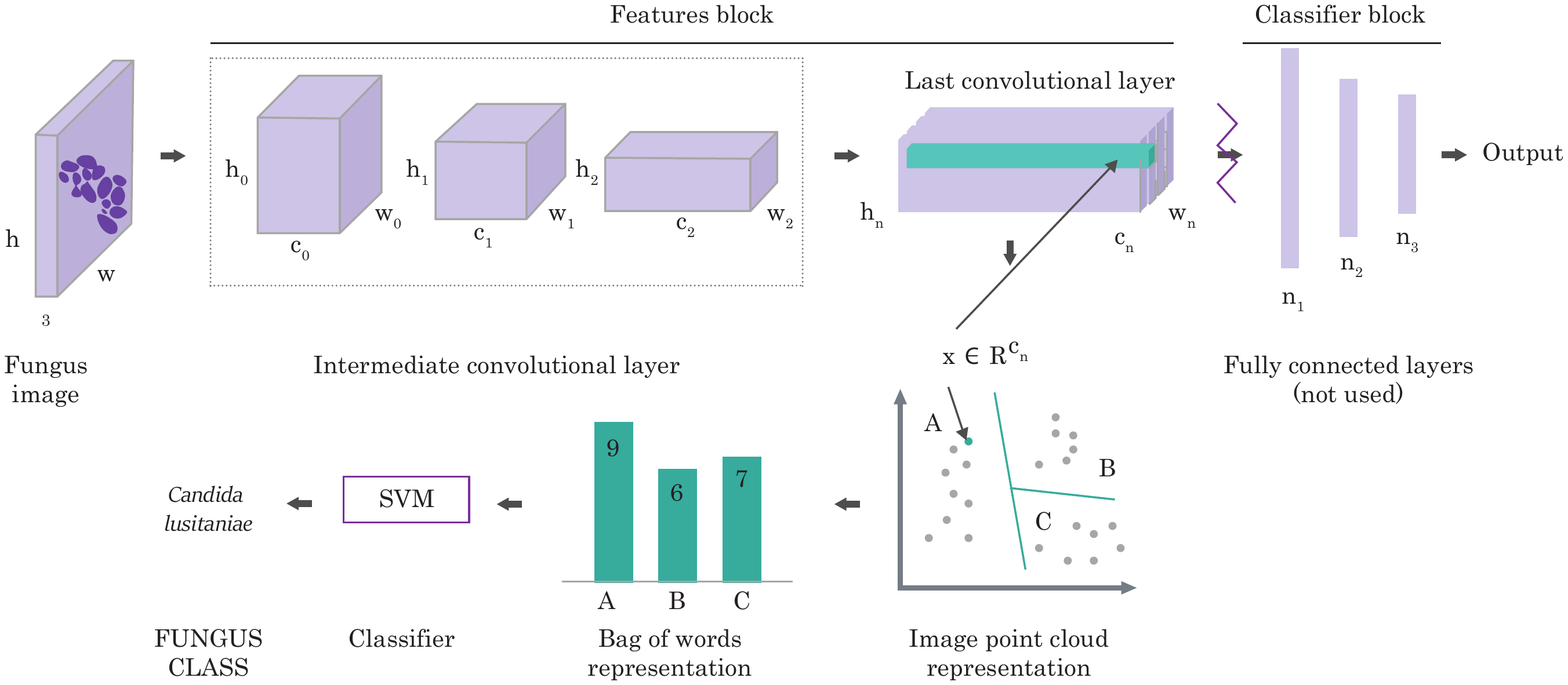}
    \caption{The deep bag-of-words multi-step algorithm produces robust image features using previously trained deep neural network, aggregates them using one of the bag-of-words approaches, and classifies them with Support Vector Machine.}
    \label{fig:algorithm}
\end{figure}

\medskip
\noindent \textbf{Method.} Deep Neural Networks (DNN) have shown human-level performance in case of large amounts of training data; however, they are limited when it comes to the application on small datasets due to the large numbers of parameters. Therefore, in this work, we consider two types of domain adaptation, both based on DNN features initially pre-trained on a different task (i.e., instance classification~\cite{deng2009imagenet}). As a baseline method, we fine-tune the classifier's block of the well-known network architectures, i.e., AlexNet~\cite{krizhevsky2012imagenet}, DenseNet169~\cite{huang2017densely}, InceptionV3~\cite{szegedy2016rethinking}, and ResNet~\cite{he2016deep} (with frozen features' block). As we present in results, such architectures are not optimal; hence, we propose to apply the deep bag-of-words multi-step algorithm shown in Fig.~\ref{fig:algorithm}. In contrast to baseline methods, which utilize ``shallow'' Neural Network to previously calculated features, our strategies aggregate those features using one of the bag-of-words approaches and then classify them with Support Vector Machine (SVM). Such a policy, previously applied to texture recognition~\cite{cimpoi2015deep} and bacteria colony classification~\cite{zielinski2017deep}, is more accurate than the baseline methods; however, it is not well known. Therefore, to make this paper self-contained, below, we describe its successive steps.

To generate robust {\bf image representation}, AlexNet~\cite{krizhevsky2012imagenet}, InceptionV3~\cite{szegedy2016rethinking} or ResNet~\cite{he2016deep} pre-trained on ImageNet~\cite{deng2009imagenet} database are used. Another option would be to use conventional handcrafted descriptors (like ORB~\cite{rublee2011orb} or DSIFT~\cite{liu2011sift}); however, they are usually outperformed by deep features. Considered network architectures consist of two parts: convolutional layers, which are responsible for extracting image features (so-called features' block), and fully connected layers, which are responsible for the classification (so-called classifier's block). Classifier's block cannot be directly used because it was trained for other types of images; however, features' block encodes more general, reusable information. Therefore, removing the classifier block from the network and preserving convolutional layers allows us to generate robust image features. In the case of AlexNet, we obtain a set of points in $256$-dimensional space, whose number depends on the input image's resolution (e.g., in case of resolution $500\times 500$ pixels, $169$ points ($13\cdot 13$) are generated).

Since the classified patches are always of the same size, their features' blocks could be used directly by the classifier. It, however, would lead to vast data dimensionality (i.e., the feature vector of size $43264$), which according to our experiments, results in the lack of generalization, primarily due to the relatively small size of the training set (100 images). Therefore, to obtain a more reliable representation of patches, we {\bf pool} the acquired set of points using Bag of Words, BoW~\cite{mccallum1998comparison,sivic2003video}, or its more expressive modification called Fisher Vector, FV~\cite{perronnin2007fisher}. The idea behind both of them is to aggregate a set of points (representing the patch) with a so-called codebook. The codebook is usually generated from the subset of training data in an unsupervised manner using a clustering algorithm (e.g., k-Means or Expectation Maximization~\cite{nasrabadi2007pattern}). Given a codebook, the set of $256$-dimensional points obtained with AlexNet for a particular image is encoded by assigning points to the nearest codeword. In traditional Bag of Words, this encoding leads to a codeword histogram, i.e., a histogram for which each codeword contains points closest to this codeword. In the case of the Fisher Vector, the clusters are replaced with a Gaussian Mixture Model (GMM), and the representation encodes the log-likelihood gradients with respect to the parameters of this model. In this paper, we will use notations {\em deep Bag of Words} and {\em deep Fisher Vector} to refer to those two types of pooling methods.

As a result of pooling, one fixed-size vector is obtained for each of the analyzed patches, which can be {\bf classified} with any machine learning methods to distinguish between various fungus species. We decided to use Support Vector Machine and Random Forest classifiers for this step.

\section*{Experimental setup and results}
\label{sec:experiments}

For the experiments, we split our DIFaS database (9 strains $\times$ 2 preparations $\times$ 10 images) into two subsets, so that both of them contain images of all strains, but from different preparation. It is because each preparation has its characteristics, and according to our previous studies~\cite{zielinski2017deep}, using images from the same preparation both in training and test set can result in overstated accuracy. As an example, let us consider the background-size, which depends on the size of the colony moved by inoculation loop from Sabouraud agar to preparation. Because there are only two preparations for each species in the dataset, the classifier could end up learning clinically irrelevant background-size instead of relevant fungus features. Therefore, images from particular preparation should not be shared between training and test set. We decided to use 2-fold cross-validation (one fold with $90$ images from the first preparations and the second fold with $90$ images from the second preparations). Moreover, we decided to classify patches instead of the whole image (see Image preprocessing for details) and introduce additional class corresponding to the background (BG) to compensate for the preparation characteristic on the final result. For each fold, we optimize the following parameters using internal 5-fold cross-validation: number of clusters in BoW $\in [5, 10, 20, 50, 100, 200, 500]$; number of clusters in FV $\in [5, 10, 20, 50]$; SVM kernel $\in [linear, RBF]$; SVM $C \in [1, 10, 100, 1000]$; SVM $\gamma \in [0.001, 0.0001]$. As the evaluation metric for grid search optimization, we use the accuracy classification score. Best results were obtained for FV with $10$ clusters and SVM with $RBF$ kernel, $C=1$, and $\gamma=0.0001$.

We performed all the experiments on a workstation with one 12 GB GPU and 256 GB RAM. On average, feature extraction, pooling, and classification take from $1$ to $2$ hours when training deep Fisher Vector. Such performance was possible thanks to the adaptation of the VLFeat library~\cite{vedaldi08vlfeat}. For comparison, the fine-tuning of the well-known architectures takes from $70$ to $85$ hours (see Table~\ref{tab:test_patch_based}). Processing time in case of baseline methods was measured by multiplying the average time of an epoch by the number of epochs till the early stopping (i.e., the increase in validation loss). In the case of deep bag-of-words approaches, processing time was computed as a sum of all three steps of the algorithm (i.e., obtaining image representation, pooling, and classification).

The remaining part of this section is structured as follows. First, we describe image preprocessing, including contrast stretching and background removal. Then, we describe the results obtained for patch-based classification using deep bag-of-words approaches and compare them with the well-known network architectures. To explain the outcomes of deep BoW, we introduce an in-depth explanatory analysis of the obtained codebooks together with the microbiological feedback. We continue this investigation for a deep FV approach. Finally, we present results obtained for scan-based classification, computed by aggregating patch-based scores. The code implemented in Python with PyTorch library is available at \url{https://github.com/bziiuj/fungus}.

\subsection*{Image preprocessing}
\label{sec:preprocessing}

\begin{figure}
    \centering
    \begin{subfigure}{0.3\textwidth}
    \includegraphics[width=\textwidth]{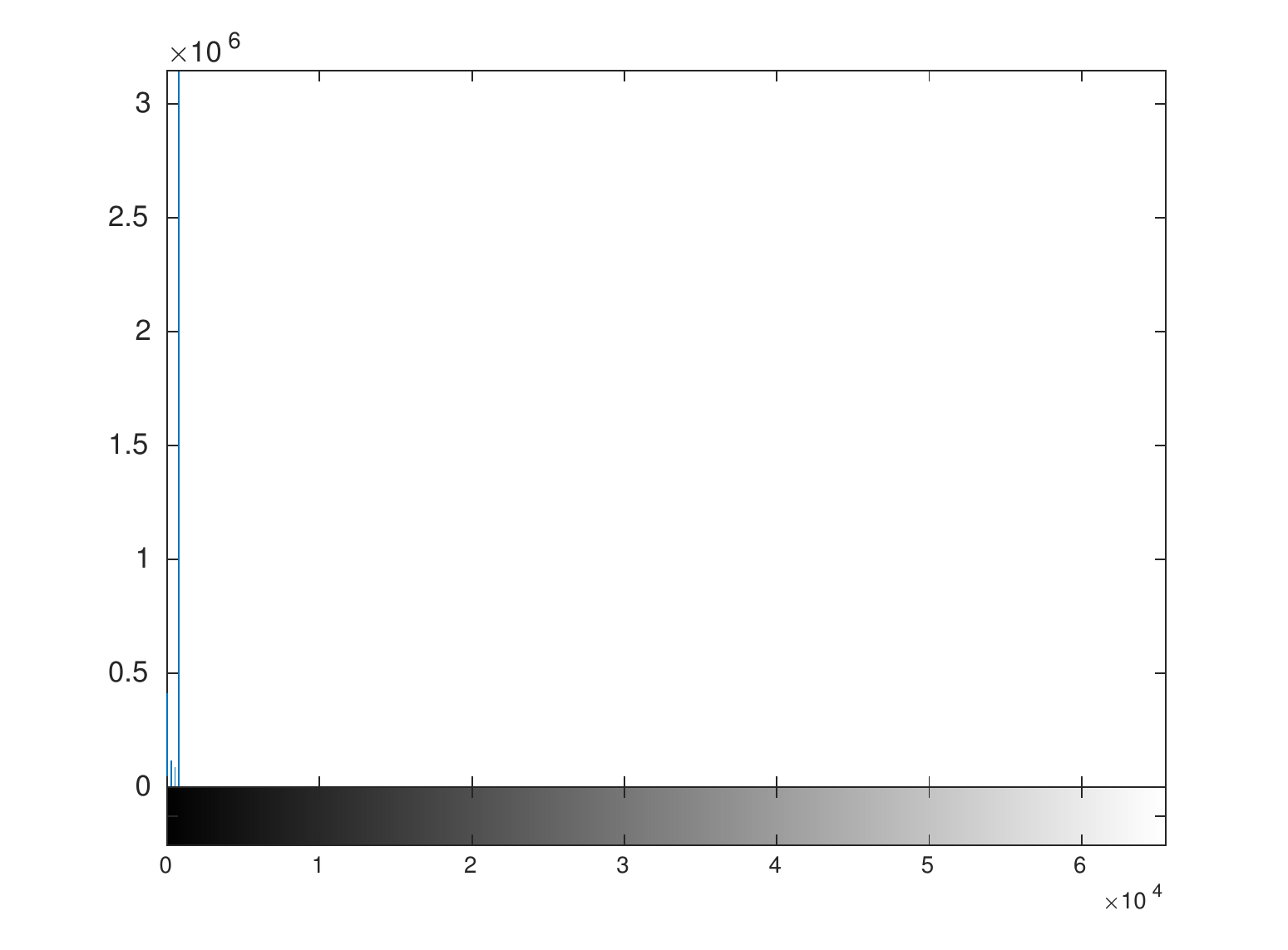}
    \caption{\footnotesize \centering Original histogram}
    \end{subfigure}
    \vspace{1em}
    \begin{subfigure}{0.3\textwidth}
    \includegraphics[width=\textwidth]{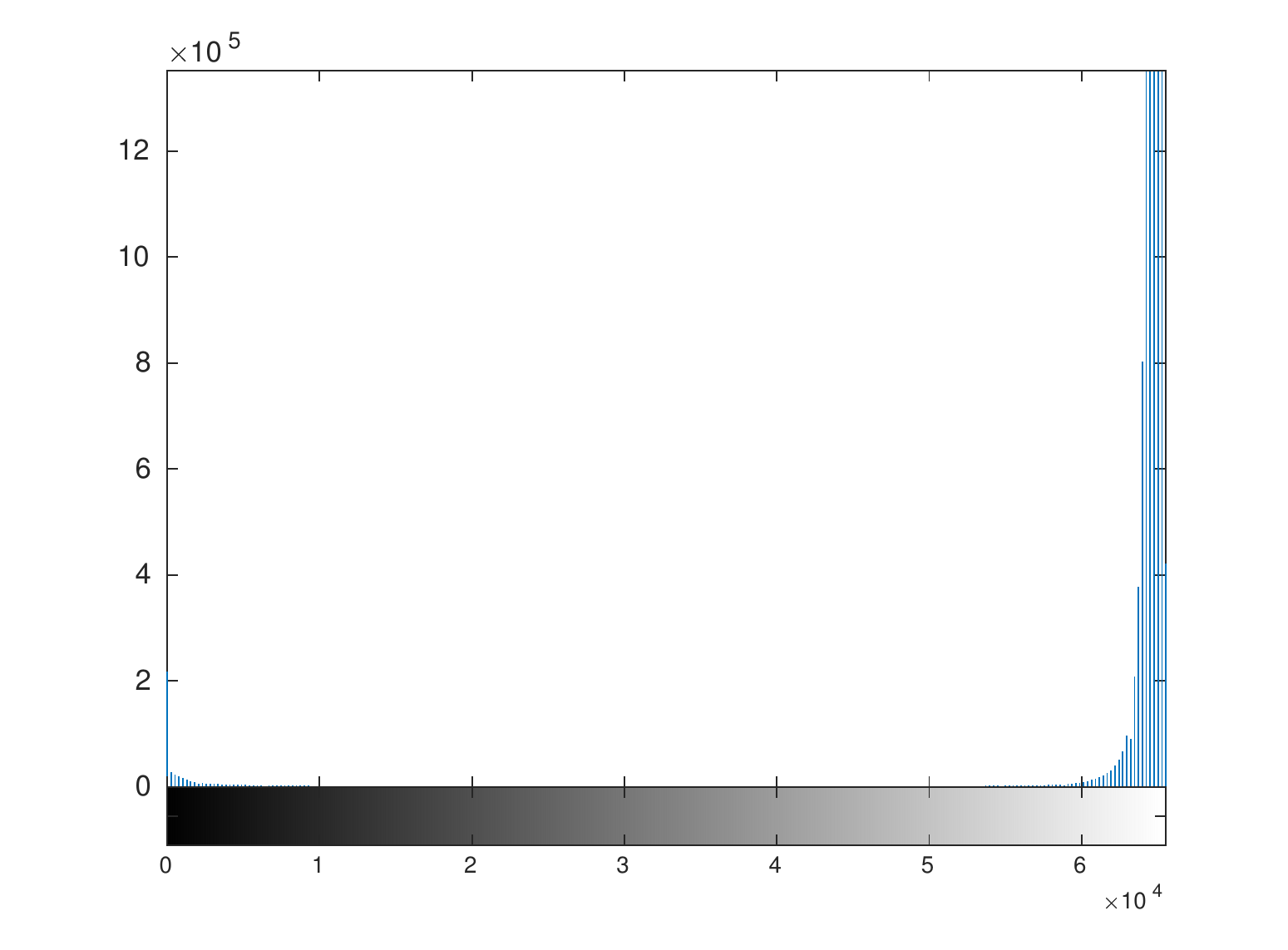}
    \caption{\footnotesize \centering Stretched histogram}
    \end{subfigure}
    \begin{subfigure}{0.3\textwidth}
    \includegraphics[width=\textwidth]{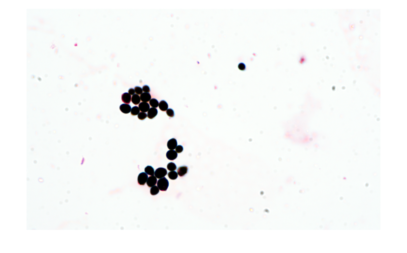}
    \caption{\footnotesize \centering Stretched image}
    \end{subfigure}
    \begin{subfigure}{0.3\textwidth}
    \includegraphics[width=\textwidth]{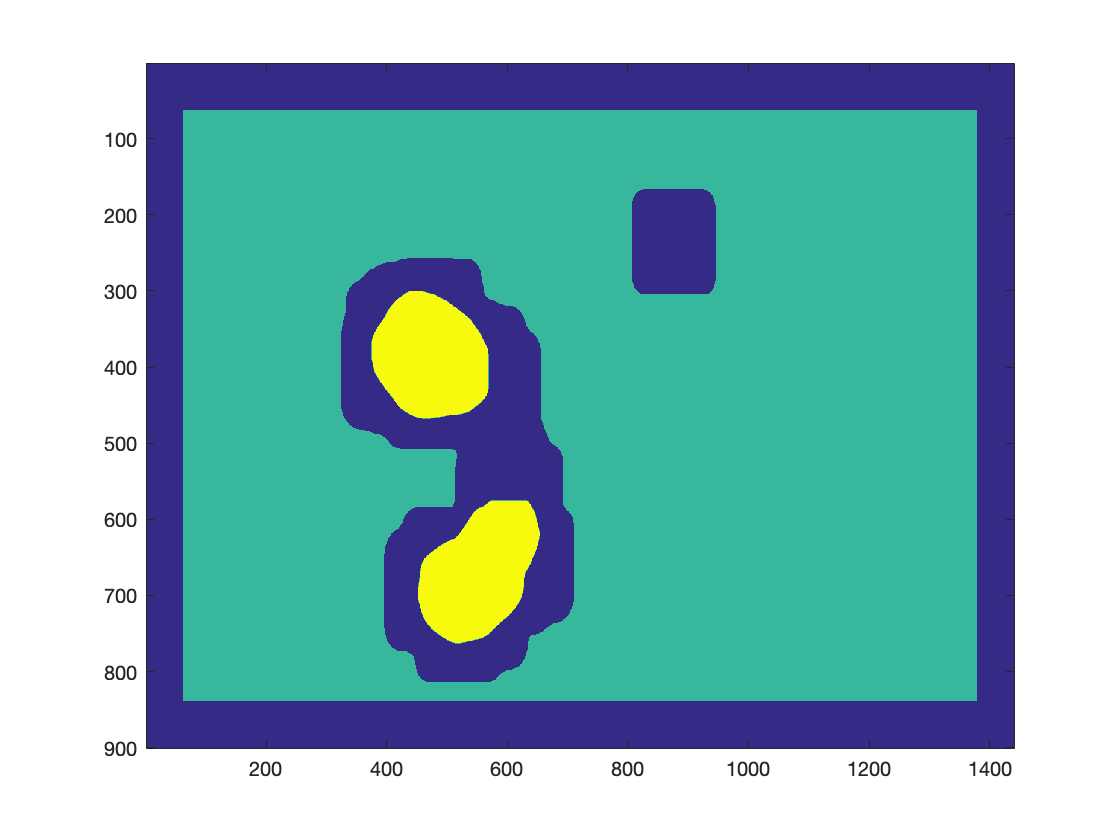}
    \caption{\footnotesize \centering FBP = 1:2}
    \end{subfigure}
    \begin{subfigure}{0.3\textwidth}
    \includegraphics[width=\textwidth]{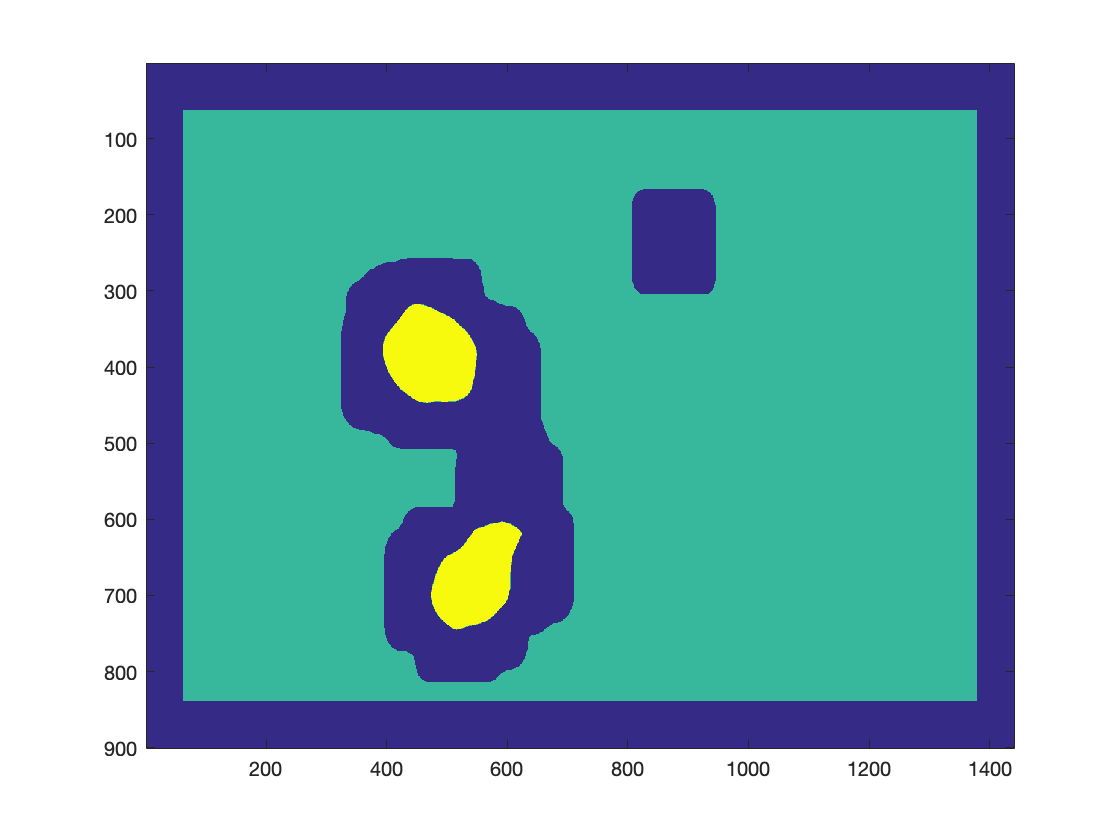}
    \caption{\footnotesize \centering FBP = 1:1}
    \end{subfigure}
    \begin{subfigure}{0.3\textwidth}
    \includegraphics[width=\textwidth]{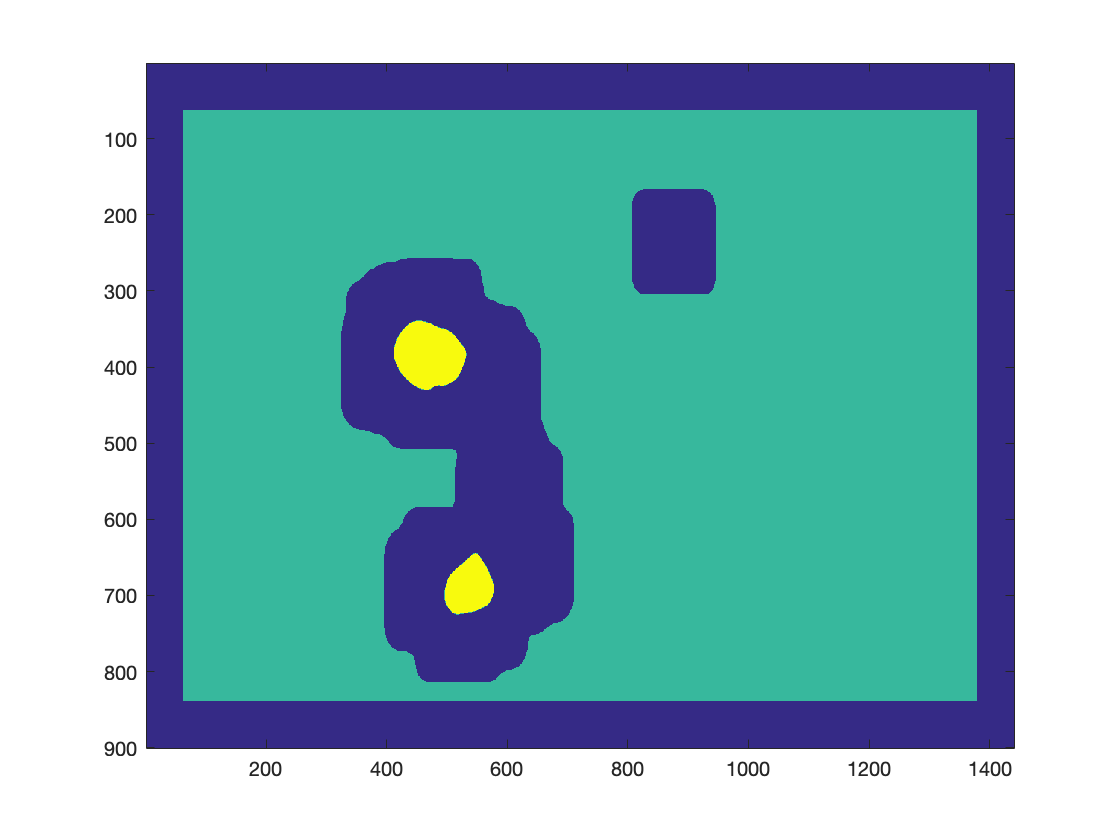}
    \caption{\footnotesize \centering FBP = 2:1}
    \end{subfigure}
    \caption{Original (a) and stretched (b) histogram of the $16$-bits image; the stretched image itself (c); and its foreground-background masks with the various foreground to background proportions (d-f). Center locations of foreground and background patches are marked as yellow and green, respectively, while blue color corresponds to the area between them (omitted during classification).}
    \label{fig:preprocessing}
\end{figure}

\noindent DIFaS database contains $180$ images of relatively high resolution and intensity range (from 0 to 65535); however, the actual pixel values are usually between 0 and 1000 (see Fig.~\ref{fig:preprocessing}a). Therefore, in the first step of preprocessing, we compute the lower and upper-intensity limits (separately for every image) and use them for contrast stretching (see Fig.~\ref{fig:preprocessing}b). Moreover, images are scaled to the range $[0, 1]$.

To overcome the issues with preparation characteristic (e.g., background-size), as the second step of preprocessing, we extract and classify only image patches with the reasonable foreground to background proportions (FBP), so patches with a rational number of foreground pixels. To obtain foreground-background segmentation on the pixel level, we apply thresholding (with threshold equal $0.5$) to a grayscaled and blurred version of the scanned image. Such a simple segmentation is sufficient and works for all the images from the dataset (see Fig.~\ref{fig:thumbnails_train} and~\ref{fig:thumbnails_test}) because the background is always much brighter than the areas with fungi cells. We tested three possible options of FBP: $1:2$, $1:1$, and $2:1$ (see Fig.~\ref{fig:preprocessing}d-f). Based on empirical studies, we decided to use FBP equal $2:1$, which gains around $1.5\%$ comparing to the other options. As a result, we obtain rough segmentations with approximated locations of foreground patches (those with FBP greater than $2:1$) and background patches (those with FBP smaller than $1:100$). Additionally, we experimented with two image scales: the original images and images scaled by factor $0.5$ (with bicubic interpolation), concluding that the latter gains around 4\% comparing to the former.

\subsection*{Patch-based classification}
\label{sec:classification}

In this experiment, we use baseline models (well-known network architectures) as well as deep Bag of Words and deep Fisher Vector models to classify each patch of the image separately. As baseline models,  we fine-tune the classifier's block of the well-known network architectures, such as AlexNet~\cite{krizhevsky2012imagenet}, DenseNet169~\cite{huang2017densely}, InceptionV3~\cite{szegedy2016rethinking}, and ResNet~\cite{he2016deep} for 100 epochs (with frozen features' block). Every baseline model was previously pre-trained on the ImageNet database~\cite{deng2009imagenet}. Before running all the experiments, we experimentally chose the optimal FBP ($2:1$), patch size ($500 \times 500$ pixels), and image scale ($0.5$) using grid search optimization. We apply data augmentation (rotations, mirror reflection, and random noise) for better regularization.

\afterpage{
\clearpage
\begin{landscape}
\begin{table}
    \centering
    \tiny
    \begin{tabular}{@{}l|rrrrrrrrrr|r|r@{}}
    \toprule
    {\bf Method} & {\bf CA} & {\bf CG} & {\bf CL} & {\bf CN} & {\bf CP} & {\bf CT} & {\bf MF} & {\bf SB} & {\bf SC} & {\bf BG} & {\bf Total} & {\bf Training time (s)} \\
    \midrule
    AlexNet     & $78.6 \pm 4.3$ & \boldmath$80.0 \pm 1.4$ & $55.8 \pm 1.4$  & $63.4 \pm 14.7$ & $75.0 \pm 7.9$ & $35.0 \pm 6.4$ & $71.6 \pm 16.6$ & $67.9 \pm 5.0$ & $72.1 \pm 2.1$  & $90.0 \pm 0.1$ & $71.6 \pm 2.4$ & $250600$ \\
    DenseNet169 & $67.9 \pm 17.9$ & $72.1 \pm 0.7$  & $53.6 \pm 0.7$ & $56.3 \pm 13.3$ & $60.0 \pm 12.7$ & $81.4 \pm 0.7$ & $68.7 \pm 5.6$ & $85.0 \pm 5.0$ & $68.8 \pm 8.6$ & $81.2 \pm 2.0$ & $72.9 \pm 0.6$ & $271600$ \\
    InceptionV3 & $67.3 \pm 1.5$ & $55.0 \pm 2.2$ & $59.3 \pm 6.4$ & \boldmath$67.0 \pm 18.2$ & $64.3 \pm 1.4$ & $81.4 \pm 2.9$ & $61.1  \pm 11.1$  & $55.0 \pm 0.7$ & $89.3 \pm 1.3$  & $84.7 \pm 3.1$ & $69.9 \pm 1.9$ & $309400$ \\
    ResNet18    & $91.4  \pm 3.6$  & $67.1 \pm 5.7$ & $67.1 \pm 5.7$ & $61.9  \pm 7.5$  & $65.0 \pm 0.7$ & $69.8 \pm 2.8$ & $54.8 \pm 12.8$  & $93.6 \pm 3.2$  & \boldmath$93.5 \pm 1.2$ & $93.1 \pm 0.5$ & $75.9 \pm 2.6$ & $263900$ \\
    ResNet50    & $86.4 \pm 3.5$ & $57.9 \pm 2.1$  & $89.3 \pm 3.5$ & $61.8 \pm 11.5$ & $60.0 \pm 7.1$ & $64.5 \pm 6.0$ & $66.4 \pm 19.3$ & $79.3 \pm 0.7$ & $64.3 \pm 12.4$  & $90.4 \pm 2.7$ & $73.9 \pm 2.6$ & $276500$ \\
    \midrule
    AlexNet BoW RF & $ 87.3 \pm 9.7 $ & $ 39.0 \pm 0.3 $ & $ 88.3 \pm 7.0 $ & $ 64.0 \pm 17.0 $ & $ 79.0 \pm 11.0 $ & $ 79.3 \pm 10.7 $ & $ 55.1 \pm 3.3 $ & $ 93.3 \pm 2.0 $ & $ 81.3 \pm 6.0 $ & $ 92.6 \pm 1.0 $ & $ 76.7 \pm 1.0 $ & $156$ \\
    InceptionV3 BoW RF & $ 44.3 \pm 7.7 $ & $ 51.0 \pm 13.7 $ & $ 60.3 \pm 0.3 $ & $ 34.2 \pm 17.8 $ & $ 28.0 \pm 0.7 $ & $ 40.7 \pm 11.3 $ & $ 13.3 \pm 1.4 $ & $ 39.7 \pm 11.7 $ & $ 36.3 \pm 8.3 $ & $ 83.5 \pm 2.0 $ & $ 44.6 \pm 0.3 $ & $155$ \\
    ResNet18 BoW RF & $ 54.3 \pm 2.3 $ & $ 39.7 \pm 8.3 $ & $ 80.7 \pm 12.7 $ & $ 46.2 \pm 20.1 $ & $ 72.0 \pm 3.3 $ & $ 61.3 \pm 11.3 $ & $ 42.1 \pm 2.9 $ & $ 58.7 \pm 8.7 $ & $ 68.0 \pm 4.0 $ & $ 91.6 \pm 2.0 $ & $ 62.7 \pm 2.4 $ & $158$ \\
    \midrule
    AlexNet BoW SVM & $ 92.3 \pm 3.0 $ & $ 44.3 \pm 18.3 $ & $ 88.7 \pm 9.3 $ & $ 62.2 \pm 22.2 $ & \boldmath$ 83.0 \pm 5.7 $ & $ 71.0 \pm 15.0 $ & $ 76.1 \pm 11.8 $ & $ 85.0 \pm 4.3 $ & $ 76.3 \pm 7.0 $ & $ 89.5 \pm 3.6 $ & $ 77.6 \pm 1.2 $ & $5124$ \\
    InceptionV3 BoW SVM & $ 44.3 \pm 10.3 $ & $ 32.3 \pm 5.7 $ & $ 57.7 \pm 2.3 $ & $ 33.3 \pm 17.6 $ & $ 28.3 \pm 1.0 $ & $ 39.7 \pm 6.3 $ & $ 13.0 \pm 1.7 $ & $ 34.7 \pm 8.0 $ & $ 32.0 \pm 2.0 $ & $ 78.8 \pm 2.1 $ & $ 40.8 \pm 1.3 $ &  $5153$ \\
    ResNet18 BoW SVM & $ 59.0 \pm 3.7 $ & $ 32.3 \pm 5.7 $ & $ 84.0 \pm 11.3 $ & $ 54.7 \pm 22.0 $ & $ 66.0 \pm 0.0 $ & $ 62.0 \pm 4.0 $ & $ 39.7 \pm 1.0 $ & $ 59.3 \pm 2.0 $ & $ 73.0 \pm 12.3 $ & $ 92.9 \pm 1.8 $ & $ 63.4 \pm 0.7 $ & $5195$ \\
    \midrule
    AlexNet FV RF & $ 83.3 \pm 10.7 $ & $ 54.3 \pm 31.7 $ & $ 81.7 \pm 0.3 $ & $ 49.3 \pm 32.0 $ & $ 78.0 \pm 12.0 $ & $ 73.0 \pm 15.7 $ & $ 76.5 \pm 5.8 $ & $ 89.3 \pm 2.0 $ & $ 74.7 \pm 6.7 $ & $ 88.0 \pm 0.9 $ & $ 75.8 \pm 0.4 $ & $166$ \\
    InceptionV3 FV RF & $ 40.7 \pm 9.3 $ & $ 53.3 \pm 11.3 $ & $ 60.3 \pm 5.0 $ & $ 37.3 \pm 21.7 $ & $ 27.7 \pm 6.3 $ & $ 47.7 \pm 13.0 $ & $ 16.5 \pm 0.2 $ & $ 35.3 \pm 13.3 $ & $ 32.0 \pm 10.7 $ & $ 81.4 \pm 4.1 $ & $ 44.6 \pm 0.7 $ & $168$ \\
    ResNet18 FV RF & $ 63.7 \pm 2.3 $ & $ 37.0 \pm 5.7 $ & $ 78.7 \pm 12.7 $ & $ 52.5 \pm 21.1 $ & $ 73.3 \pm 4.0 $ & $ 64.3 \pm 5.0 $ & $ 61.8 \pm 6.8 $ & $ 61.0 \pm 9.7 $ & $ 69.3 \pm 7.3 $ & $ 92.7 \pm 2.0 $ & $ 66.4 \pm 2.1 $ & $167$ \\
    \midrule
    AlexNet FV SVM & \boldmath$ 93.7 \pm 2.3 $ & $ 53.7 \pm 16.3 $ & \boldmath$ 90.7 \pm 4.0 $ & $ 59.6 \pm 15.2 $ & $ 77.7 \pm 14.3 $ & \boldmath$ 87.7 \pm 9.7 $ & \boldmath$ 82.8 \pm 6.9 $ & \boldmath$ 97.3 \pm 1.3 $ & $ 81.3 \pm 10.0 $ & $ 91.1 \pm 2.5 $ & \boldmath$ 82.4 \pm 0.2 $ & $1541$ \\
    InceptionV3 FV SVM & $ 46.0 \pm 14.0 $ & $ 45.0 \pm 20.3 $ & $ 58.3 \pm 3.0 $ & $ 42.2 \pm 20.4 $ & $ 24.0 \pm 5.3 $ & $ 43.7 \pm 3.7 $ & $ 13.0 \pm 1.1 $ & $ 26.7 \pm 5.0 $ & $ 76.7 \pm 3.7 $ & $ 41.3 \pm 1.9 $ & $ 41.3 \pm 1.9 $ &  $1511$ \\
    ResNet18 FV SVM & $ 71.3 \pm 11.3 $ & $ 35.3 \pm 1.3 $ & $ 59.3 \pm 10.3 $ & $ 51.6 \pm 31.1 $ & $ 77.0 \pm 8.3 $ & $ 76.7 \pm 4.7 $ & $ 57.5 \pm 0.5 $ & $ 73.3 \pm 4.7 $ & $ 77.7 \pm 9.7 $ & \boldmath$ 94.5 \pm 1.3 $ & $ 71.3 \pm 1.5 $ & $1535$ \\
    \bottomrule
    \end{tabular}
    \caption{Test accuracy of patch-based classification averaged over two runs (for two subsets described in Experimental setup).}
    \label{tab:test_patch_based}
\end{table}
\end{landscape}
\clearpage
}

\begin{figure}
    \centering
    \begin{subfigure}{0.49\textwidth}
    \includegraphics[width=\textwidth]{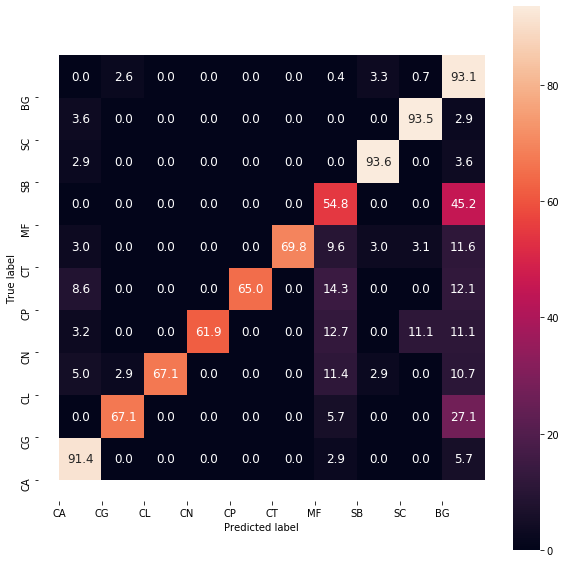}
    \caption{\footnotesize \centering ResNet18}
    \end{subfigure}
    \vspace{1em}
    \begin{subfigure}{0.49\textwidth}
    \includegraphics[width=\textwidth]{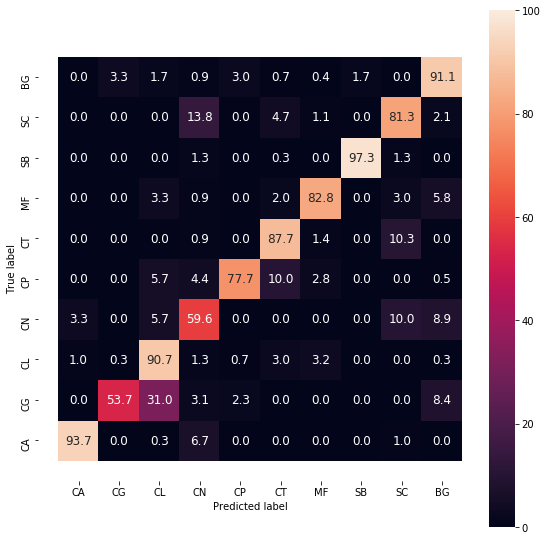}
    \caption{\footnotesize \centering deep Fisher Vector (with AlexNet and SVM)}
    \end{subfigure}
    \caption{Normalized test confusion matrices for the best baseline (ResNet18) and the best deep bag-of-words approach (deep Fisher Vector).}
    \label{fig:test_fv_confusion_matrix}
\end{figure}

The overall comparison of tested methods is presented in~Table~\ref{tab:test_patch_based}. One can observe that deep Fisher Vector works better than all the other techniques, including deep Bag of Words. However, its accuracy drops dramatically in the case of \textit{Candida glabrata} (CG) and \textit{Cryptococcus neoformans} (CN). In the case of CN it is most probably caused by a reduced number of samples, while in the case of CG due to its more substantial variance in the arrangement, appearance, and quantity (especially between two preparations, see Fig. \ref{fig:thumbnails_train} and Fig. \ref{fig:thumbnails_test}). Moreover, CG images are hard to classify due to partial discoloration (pink color instead of purple) and vast overlapping of cells. As a result, CG is often classified as \textit{Candida lustianiae} (CL) belonging to the same genus (see confusion matrix in Fig.~\ref{fig:test_fv_confusion_matrix}b). However, the classification error should decrease if the biological material of microscopic preparation has the smallest possible density with separated cells, as overlapping is the leading cause of blurriness. 

To further understand the reason for incorrect classification, we prepare a qualitative confusion matrix for deep Fisher Vector to show examples of correctly and incorrectly classified patches (see Fig. \ref{fig:fungus_cm}). We observe a high morphological similarity between misclassified species belonging to genus \textit{Candida}, \textit{Cryptococcus}, and \textit{Saccharomyces}, especially if the preparation with the biological material is discolored. Moreover, one can notice that deep Fisher Vector can return two different results for two highly overlapped patches from the same scan. It is usually caused by the artifacts in the background, such as purple trail in \textit{Candida lustianiae} (CL), predicted as \textit{Cryptococcus neoformans} (CN), or \textit{Maalasezia furfur} (MF) in Fig.~\ref{fig:fungus_cm}. The other incorrect classifications appear due to the small number of incomplete (fragmented) cells (see \textit{Candida glabrata} (CG) predicted as CN in Fig.~\ref{fig:fungus_cm}).

\afterpage{
\clearpage
\begin{landscape}
\begin{figure}
    \centering
    \includegraphics[width=1.5\textwidth]{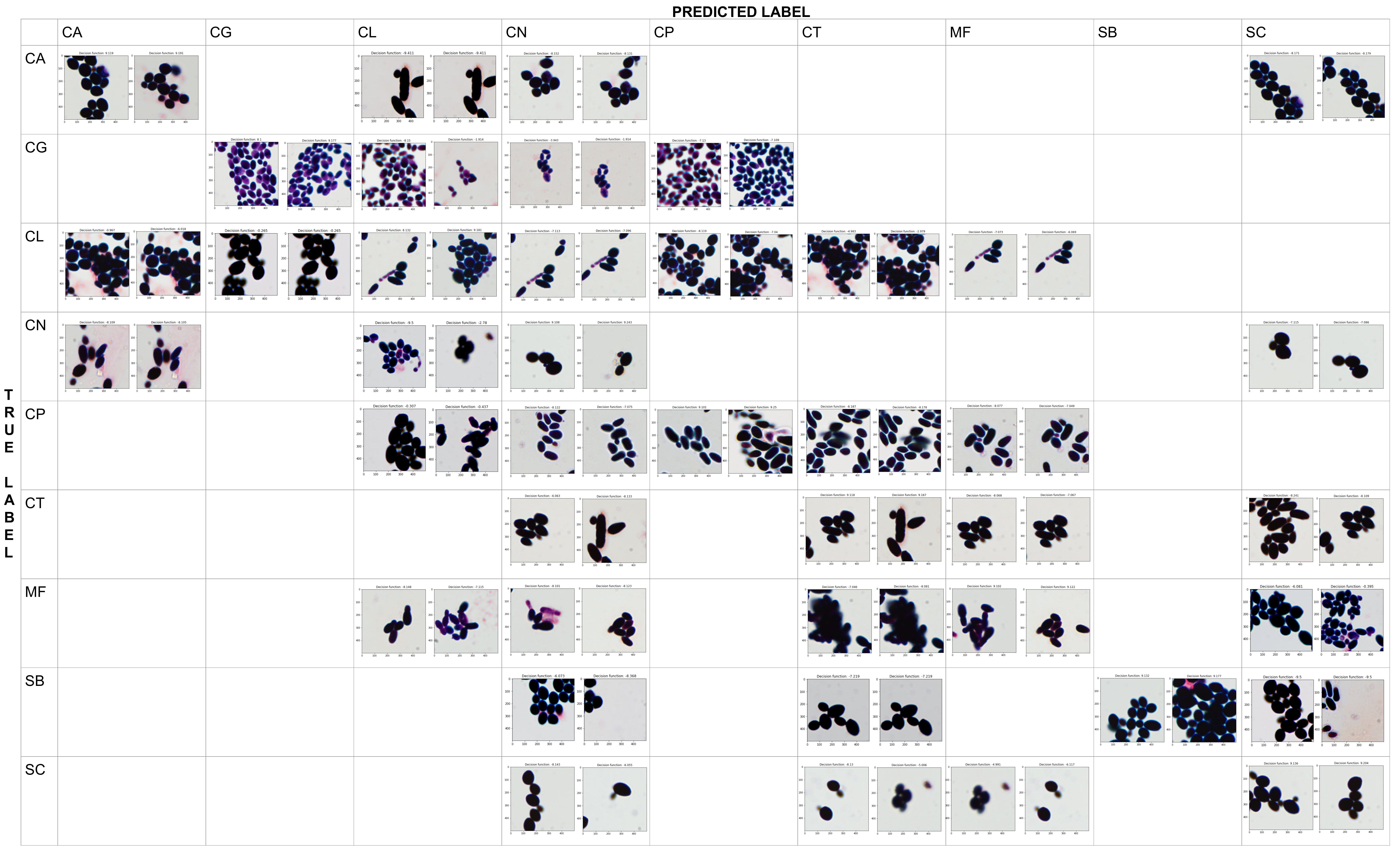}
    \caption{The qualitative confusion matrix for deep Fisher Vector (with AlexNet and SVM).}
    \label{fig:fungus_cm}
\end{figure}
\end{landscape}
\clearpage
}

\subsection*{Analysis of deep Bag of Words clusters}
\label{sec:clustersAnalysis}

In this section, we first analyze deep Bag of Words pooling step by visualizing clusters using the patches nearest to their centroids. Then, based on those patches, we introduce a description of the considered species using properties pre-defined by the microbiologists. Finally, we present the mean deep BoW for every species. To make our analysis clearer, in this section, we limit deep BoW to $10$ clusters, although its optimal number obtained with grid search optimization is $50$. Moreover, the presented properties are introduced only to explain the intrinsic rules of the method. They are not used in the automatic classification, which requires only a scan image as an input.

\begin{figure}
    \centering
    \includegraphics[width=\textwidth]{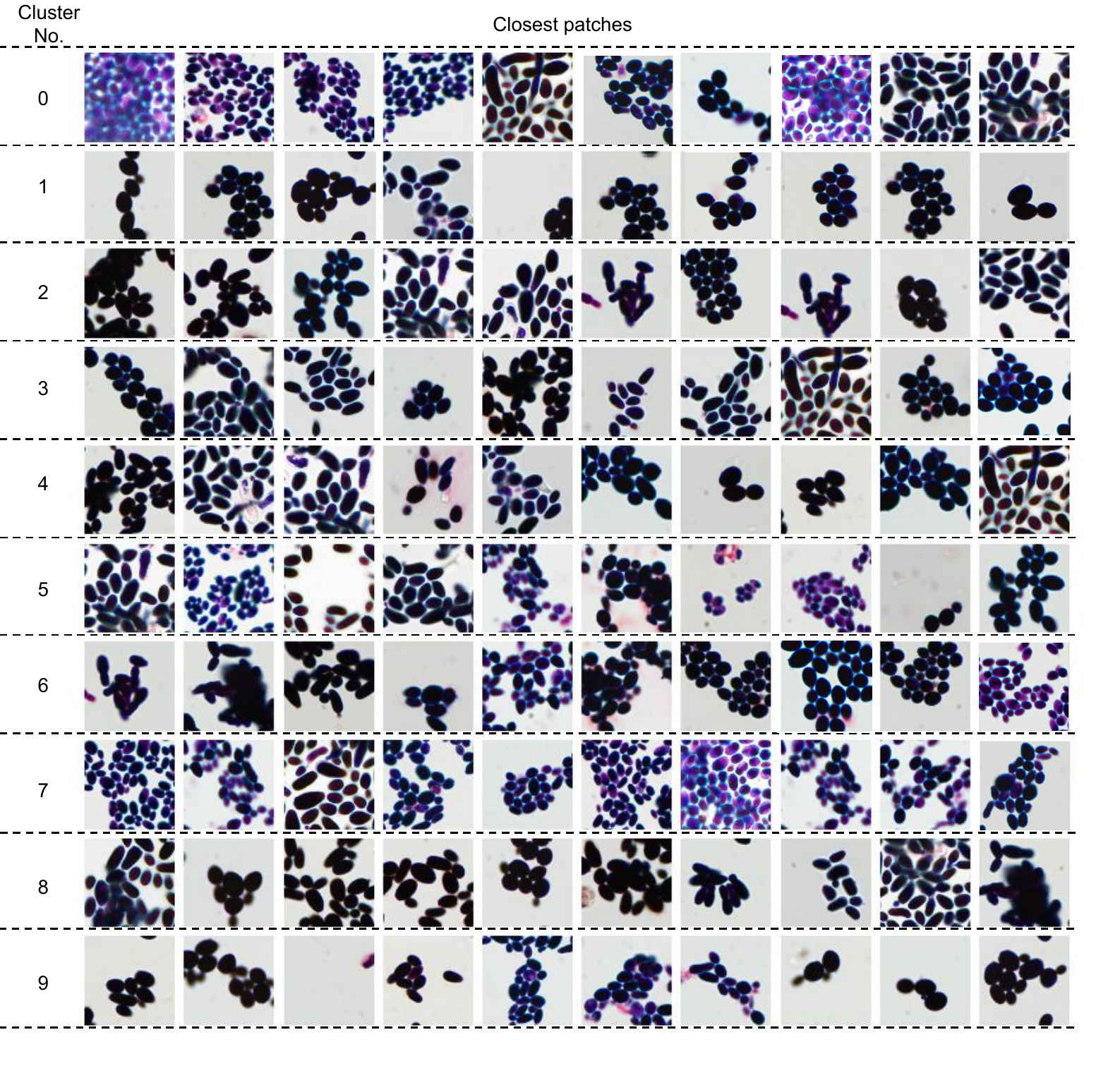}
    \caption{Ten nearest neighbors of deep Bag of Words centroids.}
    \label{fig:clusters}
\end{figure}

Ten nearest neighbors of ten deep BoW centroids obtained with k-Means algorithm are presented in Fig.~\ref{fig:clusters}. One can observe that they share common features and, therefore, can be used to determine which visual properties are essential for the classifier. We consider the following properties (see Table~\ref{tab:description}): brightness (dark or bright), size (small, medium or large), shape (circular, oval, longitudinal or variform), arrangement (regular or irregular), appearance (singular, grouped or fragmentary), color (pink, purple, blue or black), and quantity (low, medium or high). As a result, the standard set of parameters used to describe the species (size, shape, arrangement, and appearance) was significantly extended. 

\begin{table}[]
    \resizebox{\textwidth}{!}{%
    \centering
    \nohyphens{
    \begin{tabularx}{0.85\textheight}{cccccccc} \hline
    Cluster No. & Brightness & Size & Shape & Arrangement & Appearance & color & Quantity \\ [0.5ex]
    \hline\hline
    0 & bright & small & \thead{oval\\ longitudinal} & regular & \thead{grouped\\ fragmentary} & \thead{black\\ pink} & high \\
    \hline
    1 & dark & medium & \thead{oval\\ circular} & irregular & grouped & black & low \\
    \hline
    2 & dark & large & \thead{longitudinal\\ variform} & irregular & \thead{grouped\\ fragmentary} & black & medium \\
    \hline
    3 & dark & medium & \thead{variform\\ oval} & irregular & \thead{grouped\\ fragmentary} & black & medium \\
    \hline
    4 & dark & large & longitudinal & irregular & \thead{grouped\\ fragmentary} & \thead{black\\ blue} & medium \\
    \hline
    5 & bright & small & \thead{longitudinal\\ oval} & irregular & grouped & \thead{blue\\ purple} & medium \\
    \hline
    6 & dark & medium & \thead{longitudinal\\ oval} & irregular & \thead{grouped\\ fragmentary} & black & medium \\
    \hline
    7 & bright & small & \thead{longitudinal\\ oval} & \thead{irregular\\ regular} & \thead{grouped\\ fragmentary} & purple & medium \\
    \hline
    8 & dark & medium & \thead{longitudinal\\ oval} & irregular & \thead{grouped\\ fragmentary} & black & high \\
    \hline
    9 & dark & medium & oval & irregular & grouped & black & low \\[1ex]
    \hline
    \end{tabularx}
    }}
    \caption{A visual description of deep Bag of Words centroids from Fig.~\ref{fig:clusters}.}
    \label{tab:description}
\end{table}

\begin{figure}
    \centering
    \includegraphics[width=\textwidth]{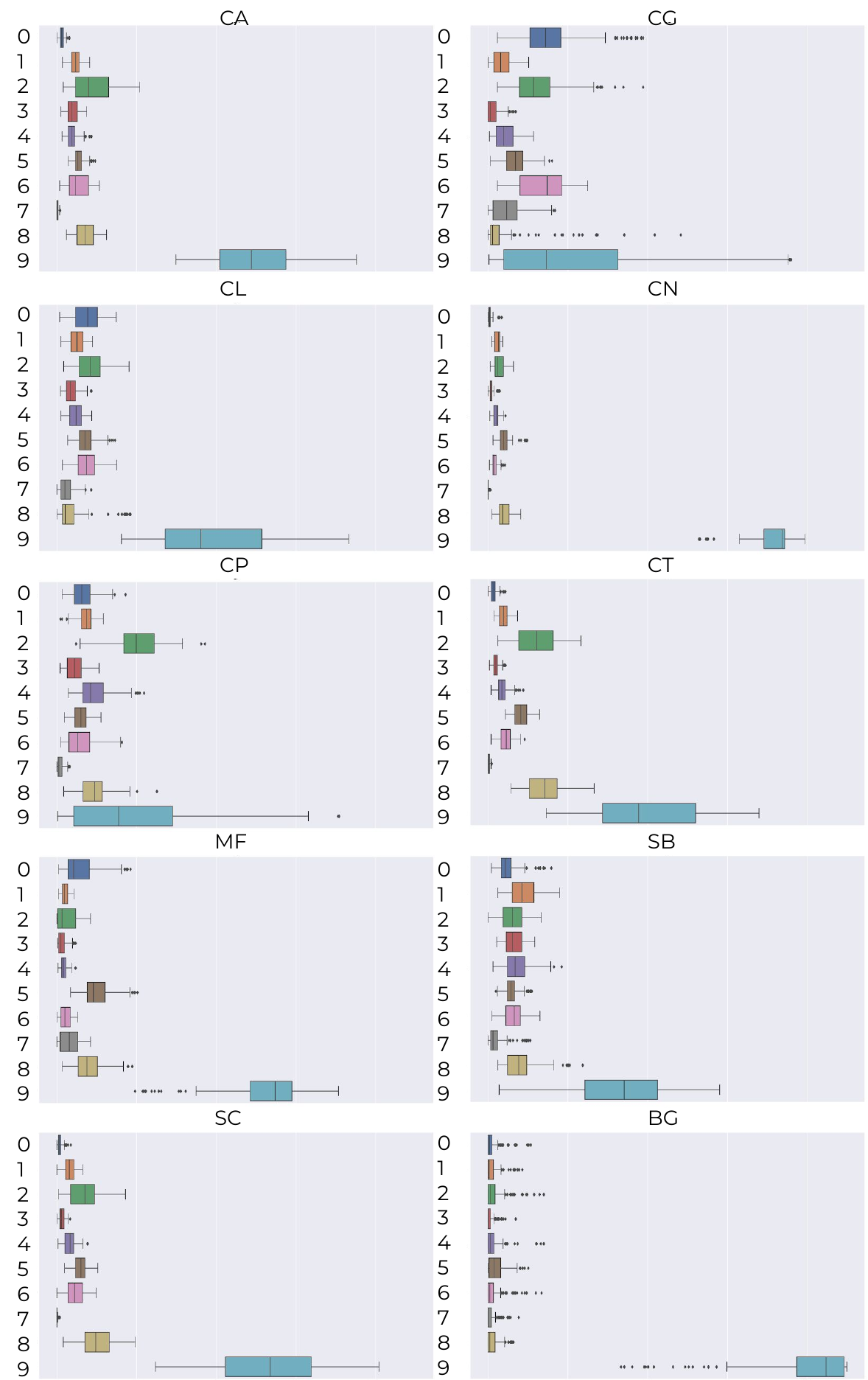}
    \caption{Mean deep Bag of Words for individual species together with the variance.}
    \label{fig:boxplot}
\end{figure}

To investigate which visual properties are essential for the classifier, we calculate mean deep Bag of Words representation for every species (see Fig.~\ref{fig:boxplot}) and then examine how the visual information about their main clusters corresponds to the knowledge of a microbiologist. The main conclusions she drew are as follows:
\begin{itemize}
    \item species of the genus \textit{Candida} mainly belong to cluster 2 with black cells of medium or large size, and oval or longitudinal shape;
    \item \textit{Maalasezia furfur} has been assigned to clusters 0, 2, 5 and 8, mostly representing the black and longitudinal shape of various size;
    \item \textit{Saccharomyces boulardii} and \textit{Saccharomyces cerevisiae} are mainly described by clusters 1, 2, 4 and 8, which are characterized by black color, medium or large size and longitudinal shape;
    \item \textit{Candida tropicalis} and \textit{Saccharomyces cerevisiae} have very similar mean Bag of Words, which confirms high morphological similarity described in~\cite{atlasGrzybow}, i.e., size $3.0$-$8.0 \times 5.0$-$10$ \si{\micro m}, oval shape, elongated, and occurring singly or in small groups).
\end{itemize}

\subsection*{Analysis of deep Fisher Vector and SVM classifier}
\label{sec:fvAnalysis}

In this section, we first analyze the power of deep Fisher Vector representation using the t-SNE algorithm~\cite{maaten2008visualizing} by projecting it on a 2D surface. Then, we analyze classifier certainty based on the scores obtained for various patches.

\begin{figure}
    \centering
    \includegraphics[width=\textwidth]{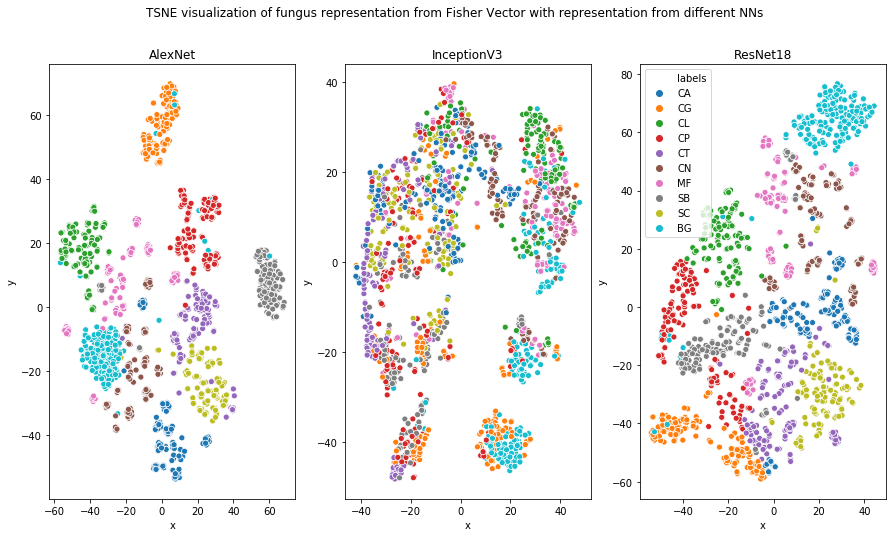}
    \caption{Projection of high-dimensional deep Fisher Vector to 2D using the t-SNE algorithm.}
    \label{fig:tsne}
\end{figure}

Projection of high-dimensional deep Fisher Vector to 2D using the t-SNE algorithm is presented in Fig.~\ref{fig:tsne}. One can observe that classes are generally well separated in the case of AlexNet and ResNet18 architectures. Nevertheless, species of the same genus are not more coherent than the other species, in contrast to what we expected. Moreover, one can observe that InceptionV3 fails domain adaptation in the case of microbiological images, which explains the results in Table~\ref{tab:test_patch_based}.

The second task in this section was to analyze classifier certainty. For this purpose, we investigate the distance of patches' representations from the classifier hyperplane, which roughly corresponds to how sure the classifier is of its decisions. Most left and right patches in Fig.~\ref{fig:svm_analysis} are correctly classified with high probability, while the ones in the middle are ambiguous. The most representative fungal \textit{Malassezia furfur} (MF) cells have oval, longitudinal shape, and often occur in the budding form, in which the daughter cells are as wide as the parent cells. While in the case of \textit{Saccharomyces cerevisae} (SC), fungal cells  characterize with round shapes, more significant in relation to \textit{Candida albicans} (CA), which are arranged individually or in small groups.

\afterpage{
\clearpage
\begin{landscape}
\begin{figure}
    \centering
    \includegraphics[width=1.5\textwidth]{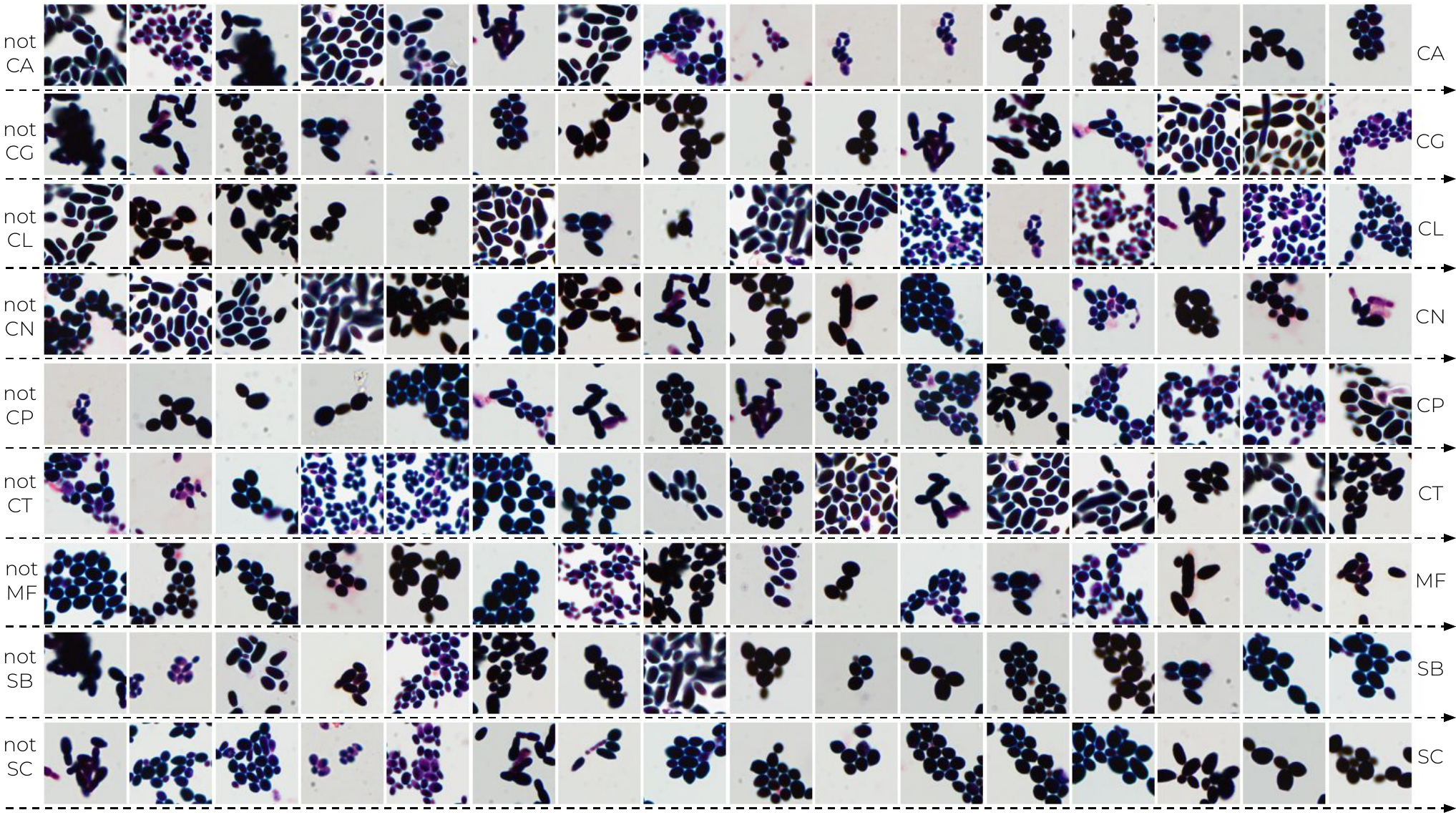}
    \caption{Classifier certainty returned together with predicted species for the deep Fisher Vector. From left to right, one can observe the most negative and most positive patches for each species.}
    \label{fig:svm_analysis}
\end{figure}
\end{landscape}
\clearpage
}

\subsection*{Scan-based classification}
\label{sec:scanClassification}

To analyze classification score for the whole scan (instead of just patches, like in previous sections), we predict classification for all foreground patches of one scan and aggregate them to obtain the most frequently predicted species. As presented in~Table~\ref{tab:test_scan_based}, deep Fisher Vector performs better than the other methods, also in this case, obtaining $15.6\%$ better accuracy than the best baseline method (ResNet18).

\afterpage{
\clearpage
\begin{landscape}
\begin{table}
    \centering
    \tiny
    \begin{tabular}{@{}l|rrrrrrrrrr|r@{}}
    \toprule
    {\bf Method} & {\bf CA} & {\bf CG} & {\bf CL} & {\bf CN} & {\bf CP} & {\bf CT} & {\bf MF} & {\bf SB} & {\bf SC} & {\bf Total} \\
    AlexNet & $95.5 \pm 5.0$ & \boldmath$95.0 \pm 5.0$ & $70.0 \pm 10.0$ & $57.0 \pm 6.9$ & $95.0 \pm 5.0$ & $45.0 \pm 10.0$ & $70.0 \pm 10.0$  & $75.0 \pm 5.0$ & $85.0 \pm 5.0$& $77.3 \pm 4.2$ \\
    DenseNet169 & $80.0 \pm 10.0$ & $85.0 \pm 5.0$  & $45.0 \pm 5.0$ & $57.0 \pm 21.2$ & $70.0 \pm 10.0$ & \boldmath$100.0 \pm 0.0$  & $85.0 \pm 15.0$ & $95.0 \pm 5.0$  & $75.0 \pm  10.0$ & $77.6 \pm 6.6$ \\
    InceptionV3 & $50.0 \pm 10.0$ & $50.0 \pm 10.0$ & $80.0 \pm 0.0$ & $78.6 \pm 14.3$ & $55.0 \pm 10.0$ & $60.0 \pm 10.0$ & $50.0  \pm 10.0$  & $85.0 \pm 5.0$  & $85.0 \pm 5.0$ & $65.9 \pm 4.9$ \\
    ResNet18 & \boldmath$100.0  \pm 0.0$  & $75.0 \pm 5.0$ & \boldmath$100.0 \pm 0.0$  & $78.6  \pm 6.9$  & $50.0 \pm 10.0$ & $70.0 \pm 0.0$ & $70.0 \pm 10.0$  & $95.0 \pm 5.0$  & $80.0 \pm 10.0$ & $78.3 \pm 5.4$ \\
    ResNet50 & \boldmath$100.0 \pm 0.0$ & $85.0  \pm 15.0$  & \boldmath$100.0 \pm 0.0$ & $57.0 \pm 6.9$ & $50.0 \pm 10.0$ & $45.0 \pm 15.0$ & $80.0 \pm 10.0$ & $95.0 \pm 5.0$  & $75.0 \pm  5.0$ & $78.1 \pm 8.3$ \\
    \midrule
    AlexNet BoW RF & \boldmath$100.0 \pm 0.0$ & $75.0 \pm 25.0$ & \boldmath$100.0 \pm 0.0$ & $75.0 \pm 15.0$ & \boldmath$100.0 \pm 0.0$ & $90.0 \pm 10.0$ & $90.0 \pm 10.0$ & \boldmath$100.0 \pm 0.0$ & \boldmath$100.0 \pm 0.0$ & $92.2  \pm  4.4$ \\
    InceptionV3 BoW RF & $80.0 \pm 20.0$ & $75.0 \pm  25.0$ & \boldmath$100.0 \pm 0.0$ & $45.0 \pm 15.0$ & $40.0 \pm 0.0$ & $60.0 \pm 30.0$ & $0.0 \pm 0.0$ & $70.0 \pm 10.0$ & $45.0 \pm 15.0$ & $57.2 \pm  2.8$ \\
    ResNet18 BoW RF & \boldmath$100.0 \pm 0.0$ & $70.0 \pm 10.0$ & \boldmath$100.0 \pm 0.0$ & $45.0 \pm 15.0$ & \boldmath$100.0 \pm 0.0$ & \boldmath$100.0 \pm 0.0$ & $80.0 \pm 10.0$ & $90.0 \pm 10.0$ & \boldmath$100.0 \pm 0.0$ & $87.2  \pm  1.7$ \\
    \midrule
    AlexNet BoW SVM & \boldmath$100.0 \pm 0.0$ & $65.0 \pm 25.0$ & \boldmath$100.0 \pm 0.0$ & $70.0 \pm 10.0$ & \boldmath$100.0 \pm 0.0$ & \boldmath$100.0 \pm 0.0$ & $85.0 \pm 5.0$ & \boldmath$100.0 \pm 0.0$ & \boldmath$100.0 \pm 0.0$ & $91.1  \pm 2.2$ \\
    InceptionV3 BoW SVM & $70.0 \pm 10.0$ & $55.0 \pm  15.0$ & \boldmath$100.0 \pm 0.0$ & $40.0 \pm 20.0$ & $45.0 \pm 5.0$ & $55.0 \pm 15.0$ & $0.0 \pm 0.0$ & $55.0 \pm 15.0$ & $40.0 \pm 20.0$ & $51.1  \pm  2.3$ \\
    ResNet18 BoW SVM & \boldmath$100.0 \pm 0.0$ & $60.0 \pm  20.0$ & \boldmath$100.0 \pm 0.0$ & $65.0 \pm 0.0$ & \boldmath$100.0 \pm 0.0$ & $90.0 \pm 10.0$ & $60.0 \pm 0.0$ & $95.0 \pm 5.0$ & \boldmath$100.0 \pm 0.0$ & $85.6  \pm  2.2$ \\
    \midrule
    AlexNet FV RF &     \boldmath$100.0 \pm 0.0$ & $65.0 \pm  35.0$ & \boldmath$100.0 \pm 0.0$ & $55.0 \pm 5.0$ & \boldmath$100.0 \pm 0.0$ & $90.0 \pm 10.0$ & \boldmath$95.0 \pm 5.0$ & \boldmath$100.0 \pm 0.0$ & \boldmath$100.0 \pm 0.0$ & $89.4 \pm  2.2$ \\
    InceptionV3 FV RF & $65.0 \pm 5.0$ & \boldmath$95.0 \pm  5.0$ & \boldmath$100.0 \pm 0.0$ & $50.0 \pm 10.0$ & $30.0 \pm 10.0$ & $75.0 \pm 25.0$ & $5.0 \pm 5.0$ & $45.0 \pm 25.0$ & $45.0 \pm 35.0$ & $56.7  \pm  3.3$ \\
    ResNet18 FV RF &    $95.0 \pm 5.0$ & $60.0 \pm  0.0$ & \boldmath$100.0 \pm 0.0$ & $65.0 \pm 5.0$ & \boldmath$100.0 \pm 0.0$ & \boldmath$100.0 \pm 0.0$ & $95.0 \pm 5.0$ & $95.0 \pm 5.0$ & $95.0 \pm 5.0$ & $89.4  \pm 1.7$ \\
    \midrule
    AlexNet FV SVM     & \boldmath$100.0 \pm 0.0$ & $75.0 \pm 25.0$ & \boldmath$100.0 \pm 0.0$ & \boldmath$75.0 \pm 15.0$ & \boldmath$100.0 \pm 0.0$ & \boldmath$100.0 \pm 0.0$ & \boldmath$95.0 \pm 5.0$ & \boldmath$100.0 \pm 0.0$ & \boldmath$100.0 \pm 0.0$ & \boldmath$93.9  \pm  3.9$ \\
    InceptionV3 FV SVM & $75.0 \pm 25.0$ & $60.0 \pm  0.0$ & \boldmath$100.0 \pm 0.0$ & $55.0 \pm 5.0$ & $45.0 \pm 15.0$ & $85.0 \pm 5.0$ & $5.0 \pm 5.0$ & $25.0 \pm 15.0$ & $45.0 \pm 25.0$ & $55.0  \pm  5.6$ \\
    ResNet18 FV SVM    & \boldmath$100.0 \pm 0.0$ & $60.0 \pm 0.0$ & \boldmath$100.0 \pm 0.0$ & \boldmath$45.0 \pm 15.0$ & $95.0 \pm 5.0$ & \boldmath$100.0 \pm 0.0$  & \boldmath$95.0 \pm 5.0$ & \boldmath$100.0 \pm 0.0$ & \boldmath$100.0 \pm 0.0$ & $88.3  \pm  2.7$ \\
    \bottomrule
    \end{tabular}
    \caption{Test accuracy of scan-based classification obtained by aggregating patch-based classification and determining the most frequent prediction.}
    \label{tab:test_scan_based}
\end{table}
\end{landscape}
\clearpage
}

\section*{Conclusions and future work}
\label{sec:conclusion}

In this paper, we apply deep neural networks and bag-of-words approaches to classify microscopic images of various fungi species. According to our experiments, the combination of features from deep neural networks with Fisher Vector works better than fine-tuning the classifier's block of the well-known network architectures and has the potential to be successfully used by microbiologists in their daily practice.

A large part of this paper is dedicated to the explainability of deep bag-of-words approaches to increase the trust in deep neural networks. For this purpose, we introduce an in-depth visual description of the properties pre-defined by the microbiologists. We hope that it will help to understand similarities and differences between fungi species better.

In our experiment, we assumed that images are obtained from the same laboratory and with the same scanner (details are presented in the Materials). However, in our opinion, this method could be easily extended to more diversified datasets by using additional preprocessing steps, which unify the input data. Due to the lack of data for such experiments, we did not cover this issue in the current article; however, it is planned for future research. Moreover, we would like to extend the DIFaS database so that it contains more preparations for all species, also gathered from other laboratories and scanners. Finally, we plan to prepare scans containing more than one species, as the automatic classification of such images would help to exclude the culture phase from the microbiological pipeline.

\section*{Acknowledgments}

This work was supported by the National Science Centre, Poland, under grants no. 2015/19/D/ST6/01215.

\bibliography{mybibfile}

\afterpage{
\clearpage
\begin{landscape}
\begin{figure}
    \centering
    \includegraphics[width=1.5\textwidth]{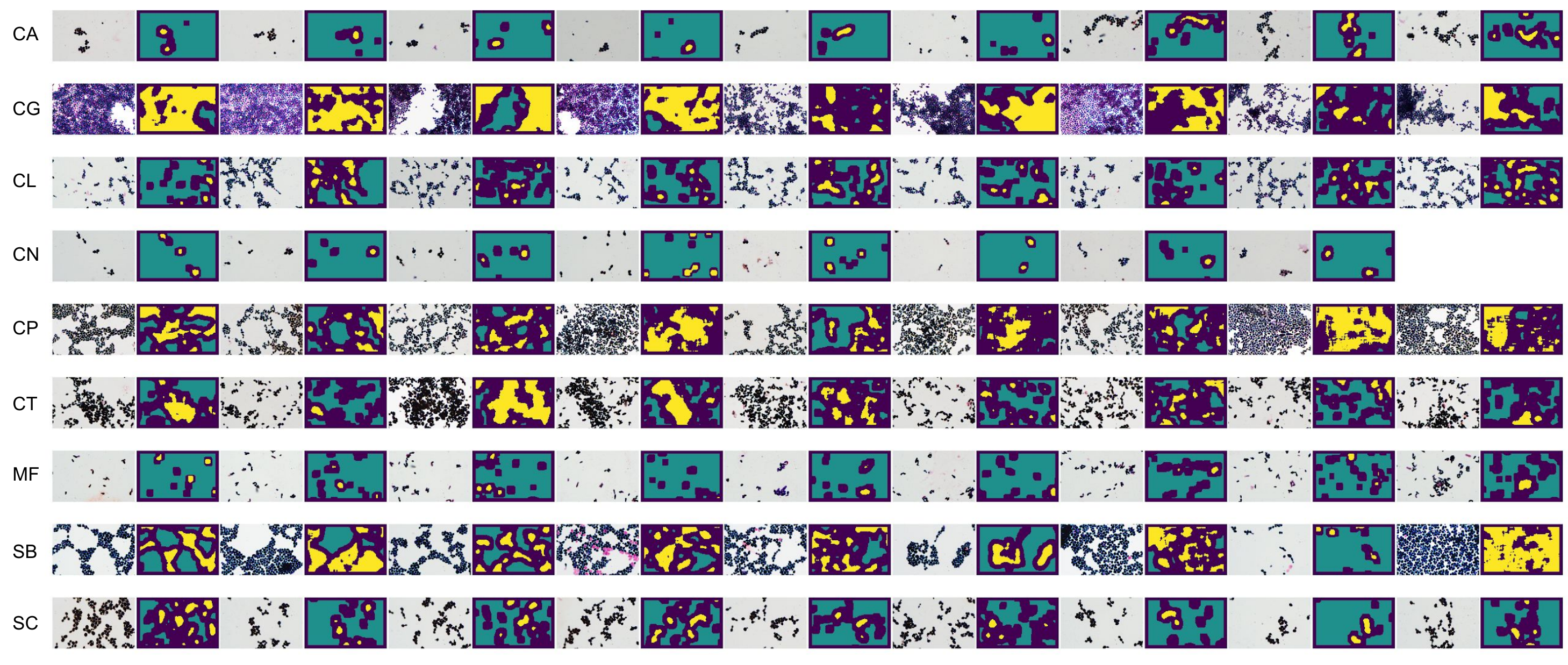}
    \caption{Images of the first preparations from the DIFaS database together with their foreground-background masks.}
    \label{fig:thumbnails_train}
\end{figure}
\end{landscape}
\clearpage
}

\afterpage{
\clearpage
\begin{landscape}
\begin{figure}
    \centering
    \includegraphics[width=1.5\textwidth]{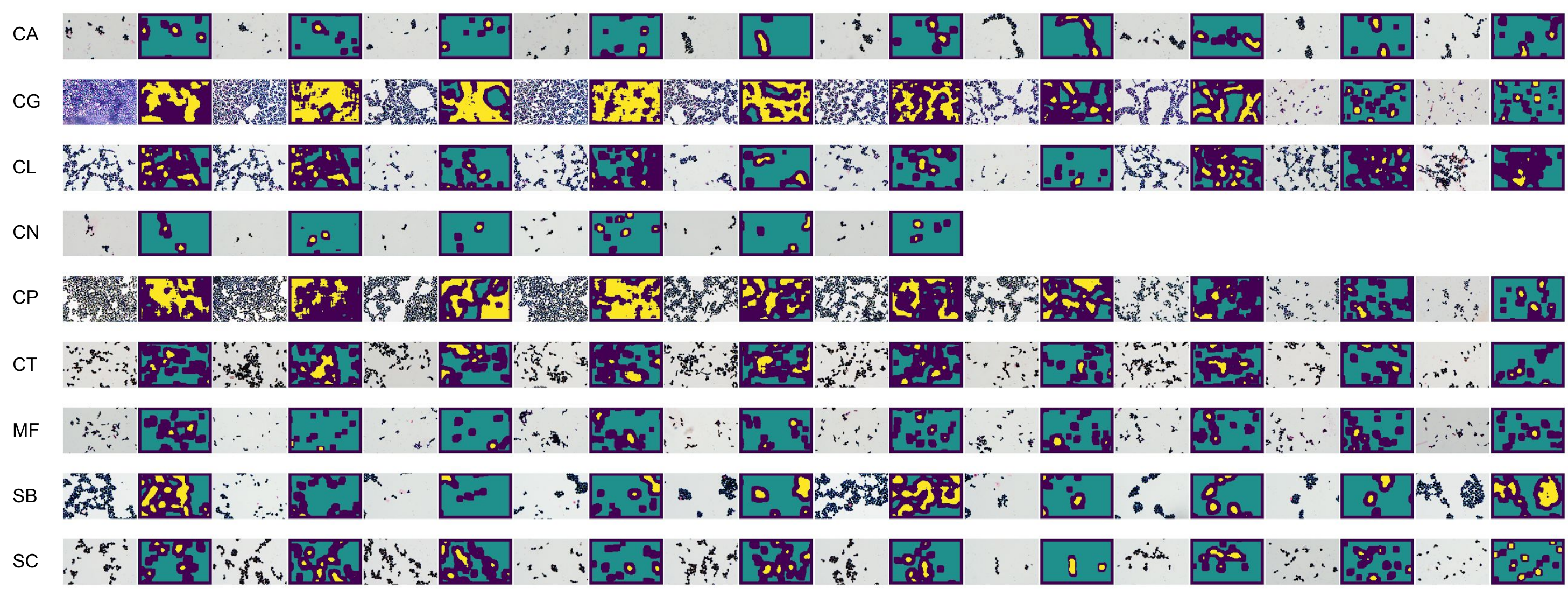}
    \caption{Images of the second preparations from the DIFaS database together with their foreground-background masks.}
    \label{fig:thumbnails_test}
\end{figure}
\end{landscape}
\clearpage
}

\end{document}